\documentclass{article}

\usepackage{PRIMEarxiv}

\usepackage[utf8]{inputenc} 
\usepackage[T1]{fontenc}    
\usepackage{hyperref}       
\usepackage{url}            
\usepackage{booktabs}       
\usepackage{amsfonts}       
\usepackage{nicefrac}       
\usepackage{microtype}      
\usepackage{lipsum}
\usepackage{fancyhdr}       
\usepackage{graphicx}       
\graphicspath{{media/}}     
\usepackage{xcolor}         
\usepackage{subfigure}
\usepackage{wrapfig}
\usepackage{amsthm}
\usepackage{amsmath}

\usepackage{algorithm}
\usepackage{algpseudocode}
\usepackage{mathtools}
\usepackage{subcaption}
\usepackage{tabularx}
\usepackage{dsfont}
\usepackage{multirow}
\usepackage{wrapfig}
\usepackage{amssymb}
\usepackage{multirow}
\usepackage{color}
\usepackage{booktabs}
\usepackage{gensymb}
\usepackage{enumitem}
\usepackage[numbers]{natbib}
\usepackage[utf8]{inputenc}
\usepackage[T1]{fontenc}

\usepackage{amsmath}
\usepackage{amssymb}
\usepackage{amsfonts}

\usepackage{graphicx}
\usepackage{subcaption}
\usepackage{booktabs}
\usepackage{multirow}
\usepackage{makecell}
\usepackage{arydshln}

\usepackage{nicefrac}
\usepackage{microtype}
\usepackage{xcolor}

\usepackage{wrapfig}
\usepackage{dblfloatfix}
\usepackage{mdframed}
\usepackage{enumitem}

\usepackage{url}
\usepackage{hyperref}

\title{Boosting Text-to-Image Diffusion Models \\ via Core Token Attention-Based Seed Selection}

\author{
Yunzhe~Zhang,~Hongfu~Liu,~and~Pengyu~Hong \\
  Brandeis University\\
  \{yunzhezhang, hongfuliu, hongpeng\}@brandeis.edu
}

\begin{document}
\maketitle

\begin{abstract}
Text-to-image diffusion models can synthesize high-quality images, yet the outcome is notoriously sensitive to the random seed: different initial seeds often yield large variations in image quality and prompt–image alignment. We revisit this “seed effect” and show that attention dynamics over prompt core tokens, the content-bearing words, measured during the first few denoising steps, strongly predict final generation quality. Building on this observation, we introduce {Attention-Based Seed Selection (ABSS)}, a training-free, plug-and-play method that ranks seeds for a given prompt by leveraging cross-attention to core tokens during the denoising process. ABSS requires no finetuning and does not alter the initial noise; it scores and ranks all candidate seeds, keeps only the top-k for full generation, and discards the rest, without relying on a fixed accept/reject threshold. Operating purely at inference time, ABSS can serve as a lightweight pre-selection add-on for existing seed-optimization pipelines, enabling additional gains. Across three benchmarks, extensive experiments show that ABSS enables consistent improvements in text–image alignment and visual quality for Stable Diffusion variants, as corroborated by human preference and alignment metrics.
\end{abstract}

\section{Introduction}
Text-to-Image Synthesis (T2I) aims to produce realistic, high-fidelity, and semantically consistent images directly from natural language prompts. Early approaches widely leveraged text-conditioned generative adversarial networks~\citep{reed2016gansynth,zhang2017stackgan,xu2018attngan}, conditional variational autoencoders~\citep{sohn2015cvae}, and autoregressive models~\citep{ramesh2021dalle} to improve semantic alignment and diversity of generated images. Diffusion models~\citep{ho2020ddpm,song2021scorebased,liu2024alignment} have driven a paradigm shift in generative modeling, establishing themselves as the leading approach through their stable training process and exceptional output quality. In particular, combining diffusion models with large-scale language/vision–language representations has driven major breakthroughs in T2I synthesis \citep{nichol2022glide,ramesh2022dalle2,saharia2022photorealistic,rombach2022ldm,balaji2022ediffi}.

\begin{figure*}[t]
    \centering
    \includegraphics[width=\textwidth]{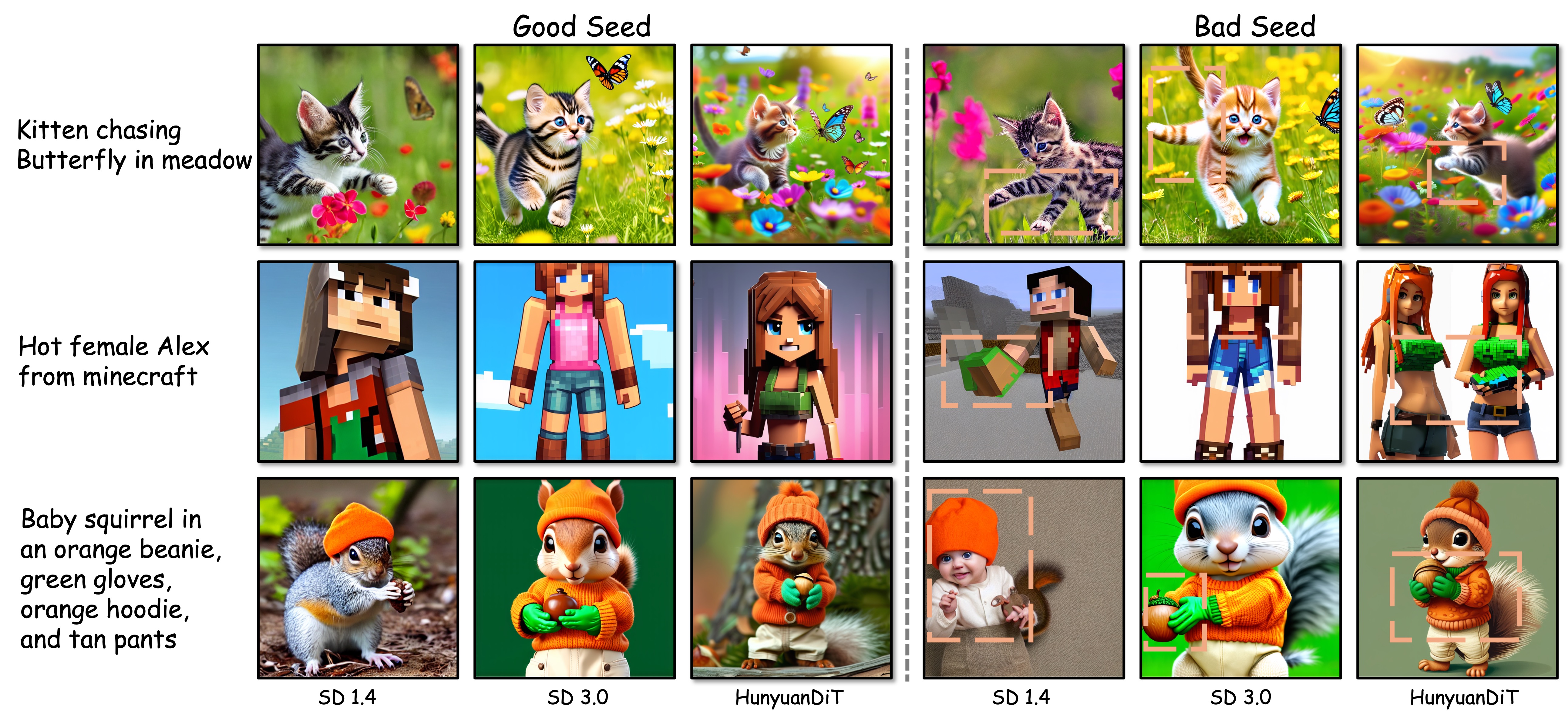}
    \caption{Illustrative examples of generations from good and bad seeds across both earlier and recent diffusion models. In earlier models such as Stable Diffusion 1.4, the gap between good and bad seeds is visually apparent, and errors from poor seeds are easily recognizable. In contrast, recent models such as Hunyuan DiT and Stable Diffusion 3.0 produce significantly higher-quality images, making outputs from bad seeds appear visually convincing at first glance. However, this improvement can obscure undesirable artifacts and semantic inconsistencies (highlighted in orange rectangles), which only become noticeable upon closer inspection. Since such errors are easier to overlook, they carry a heightened risk of propagating misleading (sometimes even harmful) effects.}
    \vspace{-4mm}
    \label{fig:motivation}
\end{figure*}

Despite recent and promising advances, diffusion-based T2I systems remain highly sensitive to the seed-defined initial latent, often yielding noticeably variable and unstable results. We refer to initial latents that reliably lead to high-fidelity, prompt-aligned, compositionally grounded, and globally coherent images as good seeds, and those that fail to do so as bad seeds. Even small perturbations to the seed initialization can significantly affect output fidelity, aesthetics, and semantic alignment~\citep{mao2023seedBrittle,li2025enhancing}. This problem is particularly evident in earlier diffusion models, but it remains far from solved in modern DiT-based~\citep{peebles2022dit} and rectified-flow T2I models~\citep{stabilityai2024sd3paper}. Although recent models substantially improve visual realism and make many generations appear visually appealing **under standard qualitative evaluation settings**, bad seeds can still produce images that fail under closer inspection, with errors in object structure, spatial layout, attribute binding, or prompt faithfulness (see Figure~\ref{fig:motivation}).

To mitigate this issue, prior work modifies the initial random latent determined by the seed to improve generation quality. We refer to studies in this line of research as seed optimization. Approaches include reward-based optimization with preference-guided objectives~\citep{eyring2024reno,miao2024noisediffusion}, attention-guided refinement via cross-/self-attention control~\citep{hong2023sag,AEchef,guo2024}, and controllable rollback techniques using inversion-based backtracking~\citep{bai2024zigzag,mao2025ctrlz, qi2024dnso}. In contrast, seed selection operates over a large pool of random seeds and focuses on identifying a subset that produces consistently high-quality outputs. \citet{goodseed2024} introduced the concept of "Golden Seeds" by running extensive generation trials on a validation set and selecting seeds that consistently yield superior results, which are then applied to the target set. Similarly, \citet{li2025enhancing} select high-performing seeds from generated datasets produced by a frozen diffusion model and subsequently reuse them for fine-tuning a new diffusion model.

\textbf{Motivations}. The motivation of this paper stems from two key aspects. First, seeds play a substantial role in shaping the quality of outputs. Second, from a practical standpoint, a single text prompt typically requires multiple generated outputs for users to evaluate and compare. Consequently, there is a strong demand for a pool of good seeds, rather than a single seed, that can consistently produce high-quality images. Unlike prior studies on seed optimization or seed-aware training, this work focuses on the \textbf{seed selection} problem: given a collection of random seeds, how can we identify those that are likely to generate high-quality outputs?

\textbf{Contribution}. In this paper, we focus on the seed selection problem. Unlike existing seed selection methods that rely on auxiliary datasets, which incurs additional seed curation/fine-tuning costs, our approach distinguishes between good and bad seeds during the denoising process itself, particularly in the early stages. Our key contributions can be summarized as follows:

\begin{itemize}[wide=10pt, leftmargin=*, nosep]
    \item We reveal a key insight: early-stage cross-attention on core tokens is a strong predictor of final prompt alignment and image quality, providing the base for deriving a simple yet effective criterion for screening and early stopping of low-quality seeds in practical inference pipelines. \vspace{2mm}
    
    \item Based on the above insight, we propose ABSS (Attention-Based Seed Selection), a training-free and plug-and-play procedure for selecting seeds at inference time, leveraging the denoising process itself without requiring any external supervision or additional model training. \vspace{2mm}

    \item We validate ABSS through extensive experiments across three benchmarks and multiple Stable Diffusion variants, and further show its effectiveness and generalization when extended to modern Diffusion Transformer based rectified-flow models with substantially different architectures.
\end{itemize}

\section{Related Work}
\textbf{Text-to-Image Diffusion Models}.
T2I diffusion models have rapidly become one of the most powerful generative model families, capable of synthesizing highly diverse and photorealistic images conditioned on natural language descriptions \citep{saharia2022photorealistic, rombach2022ldm}. These models typically incorporate pretrained language encoders---such as CLIP \citep{radford2021learning}, T5 \citep{raffel2020exploring}, or more recently large language models~\citep{balaji2022ediffi}---to transform textual inputs into dense representations. The encoded information is then injected into the generative process via cross-attention layers, which serve as the primary mechanism for aligning semantics between text and image \citep{vaswani2017attention, rombach2022ldm}. While this architecture has achieved remarkable success in controllable generation, the outputs can be highly sensitive to the initial random seeds, leading to significant quality disparity between good and bad seeds; typical failures include missing objects/parts, inappropriate overlap or misplacement, and even spurious text artifacts in inpainting \citep{goodseed2024,imagharmony2025}. Moreover, T2I models are known to suffer compositional errors—e.g., incorrect counts, positions, and attribute binding—yet recent evidence shows that a substantial portion of these failures is in fact seed-dependent \citep{li2025enhancing,gokhale2023benchmark,crystalball2025}.

\textbf{Seed Optimization in Text-to-Image Diffusion Models}. A prominent line of optimizing the seed-defined initial latent~\citep{hertz2023prompt} pursues attention-guided refinement, leveraging cross-attention maps to better align images with prompts—tokens receiving higher attention are encouraged to be more strongly expressed. Attend-and-Excite introduces a normalized attention loss that iteratively updates the latent to mitigate subject neglect \citep{AEchef}. Follow-ups further re-weight token attention or enforce coverage/consistency—without retraining—to reduce seed-specific failure modes \citep{agarwal2023,rassin2023,meral2024,guo2024,weimin2025}.
Crucially, each backward-forward cycle reconditions the trajectory with additional signals, such as refreshed attention/saliency or reward feedback, effectively increasing the informational conditioning available to the sampler and stabilizing quality across seeds. Recently, Golden Noise~\citep{zhou2024goldennoise} learns a prompt-conditioned ``noise prompt'' that transforms a random seed into a golden seed via a lightweight NPNet.

Complementary to attention guidance, Z-sampling~\citep{bai2024zigzag,mao2025ctrlz} mitigates inference-time suboptimality by inserting controlled “back” moves—partial inversion or noise re-injection—into the denoising trajectory to escape poor basins; Ctrl-Z Sampling invokes these rollbacks under a reward signal before continuing refinement. Crucially, each backward-forward cycle reconditions the trajectory with additional signals, such as refreshed attention/saliency or reward feedback, effectively increasing the informational conditioning available to the sampler and stabilizing quality across seeds. Similarly, Golden Noise~\citep{zhou2024goldennoise} learns a prompt-conditioned ``noise prompt'' that transforms a random seed into a golden seed via a lightweight NPNet.

Different from prior approaches that optimize the initial latent, in this paper we consider the seed selection problem~\citep{goodseed2024,li2025enhancing}. Specifically, we perform early-stage screening of the seed pool, discarding those that are unlikely to yield faithful generations. This selection relies on the core token of the prompt—the core semantic anchor of the description—ensuring that only seeds aligned with the main concept are preserved, leading to more reliable and efficient T2I synthesis.

\textbf{Impact of the Initial Seeds in SD}. A growing body of evidence~\citep{goodseed2024,li2025enhancing} demonstrates that the initial seed—by fixing the starting noise—substantially steers the denoising trajectory and, in turn, the final image. Seeds induce systematic biases in outputs, affecting object arrangements (typical relative placement patterns), global style and tone, subject presence or absence, and even spurious artifacts. Consequently, the same text prompt can yield consistently good or poor results depending solely on the seed. As illustrated in Figure~\ref{fig:motivation}, for the same text prompt on SD 1.4 and SD 3.0, varying seeds produce outcomes that are visibly superior or inferior. This observation has motivated a line of work~\citep{chen2023b,qu2023layoutllm,zheng2023layoutdiffusion,couairon2023zeroshot,jia2024ssmg,lian2023llmgrounded,xu2023cropdiversity,kim2026model} that tackles generation quality from the seed side—often framed as seed optimization or seed-aware guidance.

\section{ABSS:Attention-Based Seed Selection}
In this section, we present our method for seed selection in SD. A central challenge is that the quality of a seed, whether it is good or bad, can typically be assessed only after completing the entire diffusion process, which makes seed selection computationally expensive and logically problematic. To address this issue, we propose to monitor the diffusion process itself, with a particular focus on its early stages, to predict whether a given seed has the potential to generate desirable outputs. Specifically, we first analyze the evolution of attention patterns during denoising, which serve as a critical signal shaping the final image. Building on these observations, we then introduce our attention-based seed selection method for efficient inference-time screening.

\subsection{Observation During Denoising}
\looseness-1Intuitively, the entire denoising process can be viewed as a form of continuous image optimization, where each intermediate image is iteratively refined. We therefore conjecture that a successful generation should begin by establishing the main outline—capturing the core tokens that correspond to the head noun(s) denoting the primary subject(s) expected to occupy the central visual mass—followed by progressive refinement of details. By core tokens, we specifically refer to the core subject terms (e.g., \textit{cat} and \textit{dog} in “\textit{a cat and a purple dog},” or \textit{turtle} in “\textit{a fisheye lens view of a turtle in a forest}”), rather than modifiers or contextual descriptors. Appendix~\ref{app:core_tokens} provides more details on defining core tokens and further validates the robustness of \textsc{ABSS} to noisy core-token annotations. In other words, high-quality seeds should prioritize anchoring the dominant subjects before attending to secondary attributes and contextual information. By contrast, low-quality seeds disperse attention prematurely across modifiers and background tokens, which is associated with missing or undersized subjects, incorrect counts, and style-dominated artifacts.

\begin{wrapfigure}[34]{r}{0.52\textwidth} 
\vspace{-6mm}      
\centering
  \includegraphics[width=\linewidth]{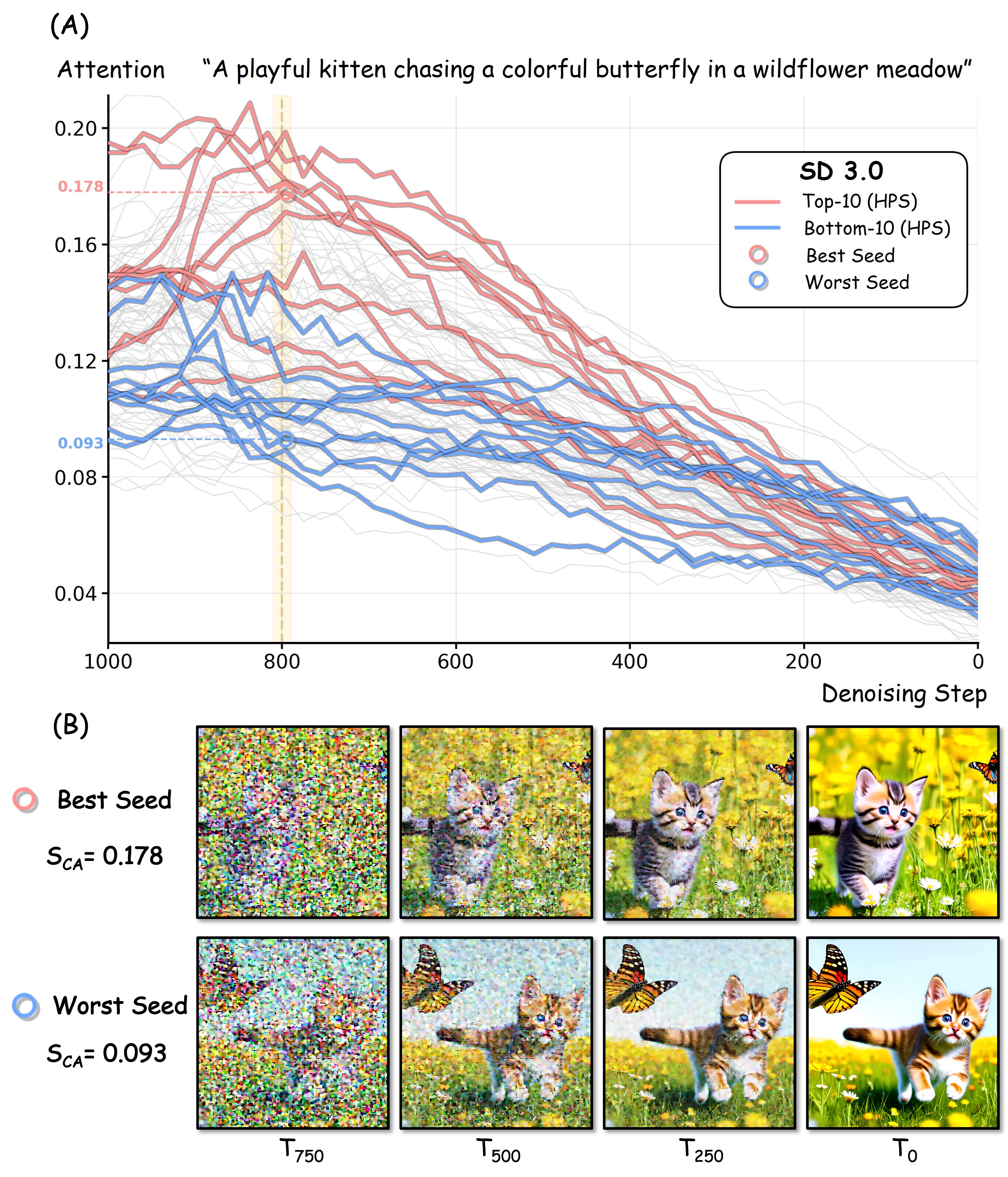} \vspace{-4mm}
  \caption{\textbf{Prompt}: ``A playful kitten chasing a colorful butterfly in a wildflower meadow.'' 
  \textbf{(A)} Trends of cross-attention on the core token ``kitten'' for 100 seeds throughout the denoising process. Red/blue curves represent high/low-quality outputs, while gray curves denote other seeds. Notably, as early as $t=800$, the cross-attention trajectories of good and bad seeds become clearly separable. 
  \textbf{(B)} Intermediate images generated from the best and worst seeds at the labeled timesteps. $S_{CA}$ denotes the cross-attention value of the core token ``kitten'' measured over the entire cross-attention map.}
  \vspace{-10mm}
  \label{fig:observation}
\end{wrapfigure}

\looseness-1 Motivated by these insights, we consider seed selection by executing only a small number of denoising steps and evaluating whether the process prioritizes the core tokens early on. To achieve this idea, we leverage cross-attention~\citep{guo2024,AEchef} on the core tokens as the criterion for early-stage seed selection. As illustrated in Figure~\ref{fig:observation}, across diverse timesteps, good seeds (red curves) consistently allocate higher attention weights to core tokens from the very beginning of the denoising trajectory, whereas bad seeds (blue curves) fail. For instance, the worst seed in Figure~\ref{fig:observation} over-attends to non-core words such as ``chasing'', which distorts the kitten legs to satisfy the action semantics.

\subsection{Method}
\begin{figure*}[!t]
    \centering
   \includegraphics[width=\textwidth]{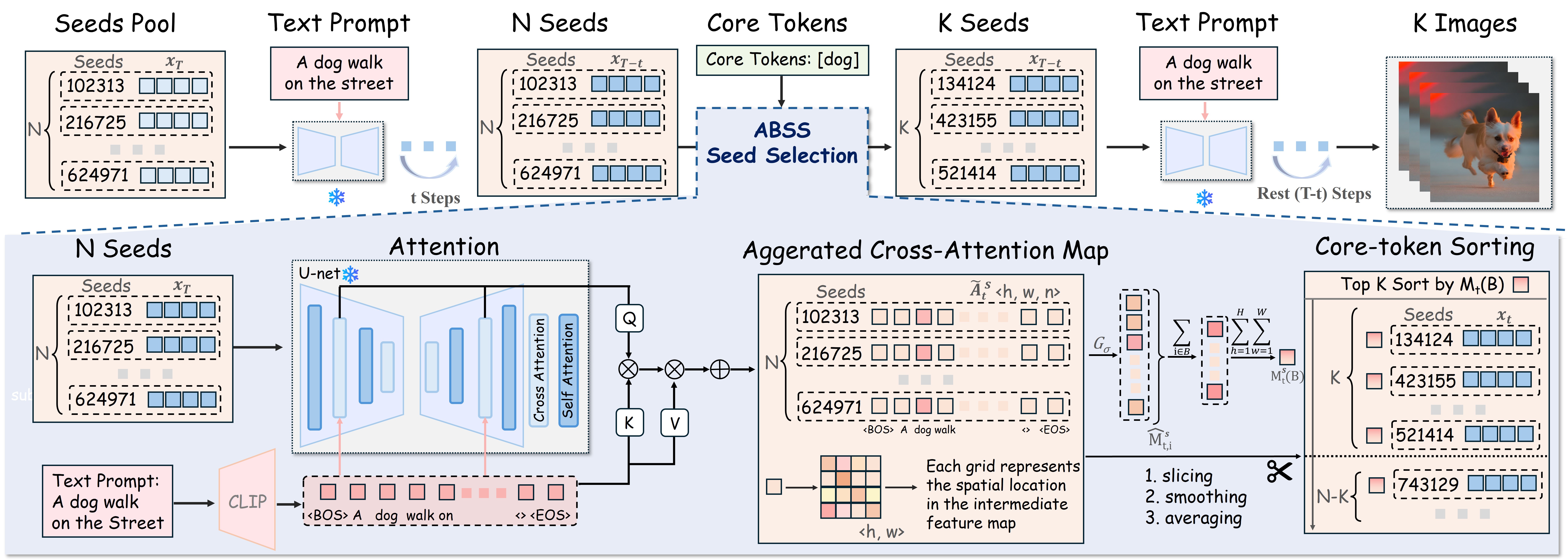}
    \caption{The Attention-Based Seed Selection (ABSS) framework. The upper pathway illustrates the standard text-to-image generation pipeline with $N$ seeds. After a few denoising steps ($t$ steps), ABSS uses cross-attention to evaluate whether the intermediate feature maps effectively capture the core tokens at an early stage. It then selects the top-$K$ intermediate feature maps to continue the denoising process.}\vspace{-4mm}
    \label{fig:method}
\end{figure*}

Figure~\ref{fig:method} illustrates the framework of our proposed Attention-Based Seed Selection (ABSS). ABSS is a lightweight and flexible seed selection method that can be seamlessly integrated into any existing SD pipeline. It identifies promising seeds by evaluating whether the intermediate images effectively capture the core tokens through the cross-attention mechanism. In the following sections, we first define the aggregated cross-attention map, and then describe how it is used to rank and select seeds during denoising process.

\textbf{Aggregated Cross-Attention Map}.
Text–image alignment in SD is achieved through cross-attention~\citep{hertz2023prompt}, which integrates textual semantics into the denoising process. A text prompt $y$ is first encoded by CLIP into a conditional embedding $c$, formulated as $c = f_{\text{CLIP}}(y)$, where $c \in \mathbb{R}^{n \times d}$ with $n$ denoting the maximum token length and $d$ the embedding dimension. The conditional embedding $c$ is then linearly projected to obtain keys $K$ and values $V$, while queries $Q$ are derived from UNet activations. For a single cross-attention layer with one head and a seed $s$ sampled from the seed pool, the attention map is computed as
\begin{equation}
\label{eq:cross-attention}
A^{s} = \mathrm{softmax}\!\left(\frac{QK^{\top}}{\sqrt{d}}\right).
\end{equation}
\looseness-1Given a specific seed $s$, the element $A^{s}_{t}[h,w,i]$ represents the attention weight assigned to token $i$ among all tokens at spatial location $(h,w)$ in the intermediate feature map at timestep $t$. Higher values indicate stronger emphasis on token $i$.

Note that the UNet architecture in SD has multiple blocks with different resolutions (commonly $64,32,16,8$), which leads cross-attention multi-head. To effectively aggregate prompt-conditioned attention, we define the {aggregated cross-attention map}: at a chosen spatial resolution with seed s, we collects cross-attention maps from the down/mid/up UNet blocks, stack all heads (and relevant blocks), reshape from the native $(\text{batch}\!\times\!\text{heads},\,q,\,n)$ format (with $q=h\cdot w$) to $(-,h,w,n)$, and average over the stacked dimension. This yields a single aggregated map $\bar{A}^{s}_{t}\in\mathbb{R}^{h\times w\times n}$. We then apply a temperature-scaled softmax along the token axis to sharpen the distribution,
\[
\tilde{A}^{s}_{t}[h,w,i]
=
\frac{\exp\!\big(\beta\,\bar{A}^{s}_{t}[h,w,i]\big)}
{\sum_{j=1}^{n}\exp\!\big(\beta\,\bar{A}^{s}_{t}[h,w,j]\big)},
\]
where $\beta$ is usually set to be 100. By this means,  for every spatial location $(h,w)$ we obtain a well-normalized probability distribution over tokens.

\paragraph{Core-token Sorting: slicing, smoothing, averaging.}
Given the aggregated attention map $\tilde{A}^{s}_{t}[h,w,i]$ for seed $s$, we focus on a subset of tokens indexed by $B$. Let $B\subseteq\{0,\ldots,n{-}1\}$ be the set of core-token indices under the original prompt indexing, including \texttt{BOS}/\texttt{EOS} (index $0$ for \texttt{BOS}, $n{-}1$ for \texttt{EOS}). For each $i\in B$, we take the spatial slice, smooth it on the $(h,w)$ grid, and average it:
\[
\begin{aligned}
M_{t,i}^{s}[h,w] = \tilde{A}^{s}_{t}[h,w,i],\ \ 
\widehat{M}_{t,i}^{s} = G_{\sigma}\ast_{\mathrm{refl}}\, M_{t,i}^{s}.
\end{aligned}
\]
\looseness-1Here $G_{\sigma}$ is a normalized $(2k{+}1)\times(2k{+}1)$ Gaussian kernel, and $\ast_{\mathrm{refl}}$ denotes 2D discrete convolution with reflection padding. Spatial pooling and averaging over core tokens gives the final core-token concentration at step $t$:
\[
M_{t}^{s}(B) \;=\; \frac{1}{|B|\,H\,W}\sum_{i\in B}\sum_{h=1}^{H}\sum_{w=1}^{W}\widehat{M}_{t,i}^{s}[h,w].
\]

Empirically, we use \(M_{t}^{s}(B)\) in ABSS to pre-screen seeds.
Consider the prompt ``\emph{A dog walk on the street}'' (illustrated in Figure~\ref{fig:method}).
Given a candidate pool \(\mathcal{S}=\{s_1,\dots,s_{N}\}\), where $N$ is the total number of seeds, we run only the first few denoising steps for each seed to obtain the aggregated cross-attention map \(\tilde{A}^{\,s}_{t}[h,w,i]\), and then compute \(M_{t}^{s}(B)\) for the core token \texttt{dog} at the last of these early steps.
We rank seeds by this score, retain the top-\(K\) to complete the remaining generation, and discard the rest. For the top-\(K\) seeds, we continue the remaining denoising steps and obtain $K$ output images. We also adapt our ABSS framework for Diffusion Transformer based architecture (See Appendix~\ref{app:dit_def}).

\begin{table*}[!t]
\centering
\caption{Mean performance on three datasets evaluated by HPS V2, ImageReward (IR), PickScore, and CLIP. Results are averaged over three generated images per prompt under each method's corresponding evaluation setting. For each metric, the best result is highlighted in bold. In the NFE column, $*$ denotes additional module training, and $\dagger$ denotes extra evaluation or gradient-based latent optimization overhead.}
\label{tab:diffusion_results_1}
\setlength{\tabcolsep}{4pt}
\resizebox{\textwidth}{!}{%
\begin{tabular}{c|c|cccc|cccc|cccc|c}
\toprule
\multirow{2}{*}{\textbf{Version}} &
\multirow{2}{*}{\textbf{Method}} &
\multicolumn{4}{|c|}{\textit{DrawBench}} &
\multicolumn{4}{c|}{\textit{InitNO}} &
\multicolumn{4}{c|}{\textit{Pick-a-Pic}} &
\multirow{2}{*}{\textbf{NFE}} \\
\cmidrule{3-6}\cmidrule{7-10}\cmidrule{11-14}
& &
\multicolumn{1}{|c}{HPS $\uparrow$} & IR $\uparrow$ & PickScore $\uparrow$ & CLIP $\uparrow$ &
HPS $\uparrow$ & IR $\uparrow$ & PickScore $\uparrow$ & CLIP $\uparrow$ &
HPS $\uparrow$ & IR $\uparrow$ & PickScore $\uparrow$ & CLIP $\uparrow$ &
\\
\midrule

\multirow{6}{*}{1.4}
  & \textsc{Random} & 0.2454 & -0.2371 & 20.6041 & 0.2552 &
            0.2711 & 0.0612 & 21.7945 & 0.2721 &
            0.2425 & -0.1964 & 20.1681 & 0.2531 & 50 \\
  & \textsc{NS} & 0.2444 & -0.2315 & 20.6192 & 0.2554 &
    0.2716 & -0.1182 & 21.7538 & 0.2754 &
    0.2436 & -0.1875 & 20.1977 & 0.2549 & 333 \\
  & \textsc{Golden} & 0.2490 & -0.1989 & 20.6341 & 0.2561 &
            0.2753 & 0.0998 & 21.8573 & 0.2720 &
            \textbf{0.2471} & -0.1852 & 20.2032 & 0.2551 & 209$^{\dagger}$ \\
  & \textsc{AE} & 0.2469 & -0.1978 & 20.4905 & \textbf{0.2569} &
            0.2724 & 0.1277 & 21.7440 & 0.2747 &
            0.2438 & -0.1984 & 20.0388 & \textbf{0.2565} & 75$^{\dagger}$ \\
  & \textsc{InitNO} & 0.2315 & -0.5618 & 20.2057 & 0.2491 &
            0.2637 & -0.2067 & 21.5729 & 0.2659 &
            0.2382 & -0.4419 & 19.8592 & 0.2489 & 100$^{\dagger}$ \\
  & \textsc{ABSS} & \textbf{0.2491} & \textbf{-0.1972} & \textbf{20.6343} & 0.2552 &
        \textbf{0.2782} & \textbf{0.1972} & \textbf{21.8881} & \textbf{0.2761} &
        0.2458 & \textbf{-0.1504} & \textbf{20.2047} & 0.2542 & 73 \\
\midrule

\multirow{5}{*}{2.1}
  & \textsc{Random} & 0.2522 & 0.0097 & 20.9931 & 0.2600 &
            0.2832 & 0.8369 & 22.2425 & 0.2915 &
            0.2634 & 0.1675 & 20.6432 & 0.2668 & 50 \\
  & \textsc{NS} & 0.2596  & 0.0539 & 21.0366 & 0.2572 &
     0.2889& 0.6919 & 22.2391 & 0.2880 &
     0.2703 & 0.1782  & 20.7239 & 0.2695 & 333 \\
  & \textsc{Golden} & 0.2643 & 0.2502 & 21.1316 & 0.2639 &
            0.2954 & 1.0196 & 22.4292 & 0.2927 &
            0.2715 & \textbf{0.3218} & 20.7782 & 0.2713 & 100$^{\dagger}$ \\
  & \textsc{ND} & \textbf{0.2663} & 0.2598 & 21.1283 & 0.2645 &
        \textbf{0.2983} & \textbf{1.2138} & 22.4425 & 0.2950 &
        0.2711 & 0.2784 & 20.7798 & 0.2731 & 550$^{\dagger}$ \\
  & \textsc{ABSS} & 0.2616 & \textbf{0.2778} & \textbf{21.1388} & \textbf{0.2660} &
            0.2972 & 1.0963 & \textbf{22.4731} & \textbf{0.2952} &
            \textbf{0.2728} & 0.2866 & \textbf{20.7842} & \textbf{0.2738} & 73 \\
\midrule
\multirow{5}{*}{\makecell{Hunyuan\\DiT}}
  & \textsc{Random} & 0.2982 & 0.8133 & 21.7436 & 0.2594 &
            0.3451 & 1.7546 & 23.7005 & 0.3058 &
            0.3162 & 1.1448 & 22.2266 & 0.2734 & 50 \\
  & \textsc{NS} & 0.2992 & 0.7814 & \textbf{21.8136} & 0.2617 &
    0.3470 & 1.7431 & 23.7872 & 0.3069 &
    0.3189 & 1.1250 & 22.2188 & 0.2751 & 333 \\
  & \textsc{Golden} & 0.2991 & 0.8026 & 21.7349 & 0.2585 &
            0.3484 & \textbf{1.7598} & 23.7366 & 0.3072 &
            \textbf{0.3198} & 1.1443 & 22.2199 & 0.2738 & 209$^{\dagger}$ \\

  & \textsc{NPNet} & 0.2955 & 0.7093 & 21.7486 & 0.2564 &
        0.3467 & 1.7561 & 23.7126 & 0.3062 &
        0.3148 & \textbf{1.1456} & 22.2203 & 0.2649 & 50$^{\dagger *}$ \\
  & \textsc{ABSS} & \textbf{0.3002} & \textbf{0.8138} & 21.7618 & \textbf{0.2628} &
        \textbf{0.3497} & 1.7591 & \textbf{23.7894} & \textbf{0.3078} &
        0.3190 & 1.1446 & \textbf{22.2284} & \textbf{0.2760} & 71 \\
\midrule     
\multirow{6}{*}{3.5-L}
  & \textsc{Random} & 0.3003 & 1.0374 & 22.2218 & 0.2822 &
            0.3358 & 1.7713 & 23.3128 & 0.2921 &
            0.3208 & 1.2031 & 22.5108 & 0.2836 & 50 \\
  & \textsc{NS} & 0.3038 & 1.0991 & 21.9421 & 0.2828 &
        0.3374 & 1.8234 & 23.3473 & 0.2988 &
        0.3221 & 1.2142 & 22.5564 & 0.2833 & 333 \\
  & \textsc{Golden} & 0.3039 & 1.1298 & 22.0218 & 0.2819 &
            0.3401 & 1.7882 & 23.3689 & 0.2935 &
            0.3249 & 1.2153 & 22.5221 & 0.2814 & 209$^{\dagger}$ \\
  & \textsc{ND} & 0.2951 & 1.0658 & 21.8575 & 0.2859 &
        0.3369 & 1.8261 & \textbf{23.6984} & \textbf{0.3048} &
        0.3231 & 1.2215 & \textbf{22.6885} & \textbf{0.2864} & 550$^{\dagger}$ \\
  & \textsc{CoRe$^2$} & 0.2993 & \textbf{1.1623} & 21.2369 & 0.2863 &
    0.3339 & 1.8041 & 23.4675 & 0.3036 &
    0.3196 & 1.1572 & 22.3304 & 0.2841 & 50$^{\dagger *}$ \\
  & \textsc{ABSS} & \textbf{0.3052} & 1.1342 & \textbf{22.2923} & \textbf{0.2869} &
            \textbf{0.3411} & \textbf{1.8297} & 23.4522 & 0.2934 &
            \textbf{0.3258} & \textbf{1.2362} & 22.5370 & 0.2834 & 71 \\
\bottomrule
\end{tabular}
}
\vspace{-6mm}
\end{table*}

\section{Empirical Analysis}
We conduct extensive experiments and evaluate the effectiveness and generality of ABSS through three guiding questions:
\textit{Q1: How does ABSS compare with existing seed selection and optimization strategies?} We compare ABSS against both seed-selection and seed/noise optimization baselines. For all methods, we evaluate the average k images generation quality. We further analyze their computational costs to assess the trade-off between generation quality and inference-time overhead.
\textit{Q2: Can ABSS serve as a plug-in to boost state-of-the-art seed or initial-noise optimization methods?} We use ABSS as a seed filter while keeping each downstream optimizer unchanged to assess additive gains.
\textit{Q3: Is focusing solely on core tokens sufficient to improve performance?} We ablate non-core tokens and test whether core-token-only yields consistent improvements without additional token-level modeling. 

\subsection{Experimental setting}
\looseness-1\textbf{Datasets}. We evaluate on three prompt suites: \textit{DrawBench}\citep{saharia2022photorealistic}, a broad suite covering compositionality, commonsense, spatial relations, and fine-grained attributes; \textit{InitNO}~\citep{guo2024}; and \textit{Pick-a-Pic} \citep{kirstain2023pickapic}, a large-scale human-preference–oriented benchmark used for alignment studies. These datasets jointly stress both semantic faithfulness and aesthetic preference, enabling a balanced assessment of text–image generation quality. 

\looseness-1\textbf{Metrics}. We report four metrics. Three are preference-oriented: HPS (v2.1)~\citep{wu2023hpsv2}, ImageReward (IR)~\citep{xu2024imagereward}, and PickScore~\citep{kirstain2023pickapic}—each trained on large-scale human-preference data and shown to correlate well with human judgments. We also include CLIP~\citep{cherti2023reproducible, schuhmann2022laion5b, radford2021learning} text--image similarity as a proxy for semantic alignment between the prompt and the generated image. Higher values indicate better performance for all reported metrics. To measure computational cost, we report both the effective number of function evaluations (NFE) and GPU latency. We define NFE as the number of full denoising-model forward passes, where one complete pass through all DiT blocks or U-Net layers counts as one NFE. Thus, DDPM/DDIM inversion steps are included in NFE whenever they invoke the full denoising model. GPU latency measures the actual wall-clock inference time, including online overheads such as latent optimization via backpropagation, auxiliary-module execution, evaluation, and seed/noise screening.

\looseness-1\textbf{Implementation Details}. We used eight text-to-image backbones, covering both classical U-Net diffusion models and recent DiT/rectified-flow architectures: Stable Diffusion v1.4/1.5~\citep{rombach2022ldm},
Stable Diffusion v2.0/2.1~\citep{sd20release,sd21release},
Stable Diffusion v3.0 Medium, Stable Diffusion v3.5 Large~\citep{stabilityai2024sd3paper}, FLUX.1~\citep{flux1devcard,greenberg2025demystifying},
and Hunyuan-DiT~\citep{li2024hunyuandit}. SD~1.4 is a widely used 512-resolution baseline; SD~1.5 improves prompt following and aesthetics; SD~2.0 updates the text encoder and supports higher resolutions; SD~2.1 further improves alignment and detail. Beyond diffusion-based U-Net backbones, we also extend \textsc{ABSS} to DiT-based and rectified-flow backbones, including SD~3.0 Medium, SD~3.5 Large, FLUX.1, and Hunyuan-DiT. SD~3.0 Medium and SD~3.5 Large introduce stronger Multimodal Diffusion Transformer backbones with improved text-image alignment and visual quality. FLUX.1 and Hunyuan-DiT further represent recent large-scale rectified-flow/DiT-based models, allowing us to test \textsc{ABSS} on modern non-U-Net architectures. For each backbone and prompt, \textsc{ABSS} samples a seed pool of 10, evaluates $M_{t}^{s}(B)$ at timestep $t{=}10$ with total 50 inference steps, and retains top-$k{=}3$; metrics are averaged over the 3 generated images. We keep the seed-pool size, step budget, and evaluation protocol consistent across backbones whenever applicable, with only model-specific settings such as native resolution, scheduler, and attention-map extraction adjusted accordingly. We provide full implementation and sampling details, including scheduler, guidance scale, step budget, resolution, and attention aggregation strategy, in Appendix~\ref{sec:supp_table1_setting}.


\subsection{Comparison with seed selection and optimization strategies (Q1)}

\looseness-1We compare with a total of 8 baseline methods to answer Q1, covering both seed-selection and optimization strategies. \textsc{Random}, \textsc{Golden} Seeds (\textsc{Golden}) \citep{goodseed2024}, and Noise Selection (\textsc{NS}) \citep{qi2024dnso} are used as baselines because they can be applied across different T2I backbones. \textsc{Random} samples distinct seeds uniformly for each prompt, \textsc{Golden} selects a small validation split of prompts for each dataset, ranks candidate seeds by their average validation performance, and then uses the top-ranked seeds for the remaining prompts. \textsc{NS} selects seed according to a stability criterion instead of random sampling. In the main paper, we use four representative backbones for the primary comparison: SD~1.4, SD~2.1, SD~3.5 Large, and Hunyuan-DiT. Results on the remaining backbones with baseline methods, including SD~1.5, SD~2.0, SD~3.0 Medium, and FLUX.1, are provided in Appendix  \ref{sec:results_q1_table}.

\begin{figure*}[!t]
    \centering
    \vspace{-2mm}

    \includegraphics[width=\linewidth, height=0.66\textheight, keepaspectratio]{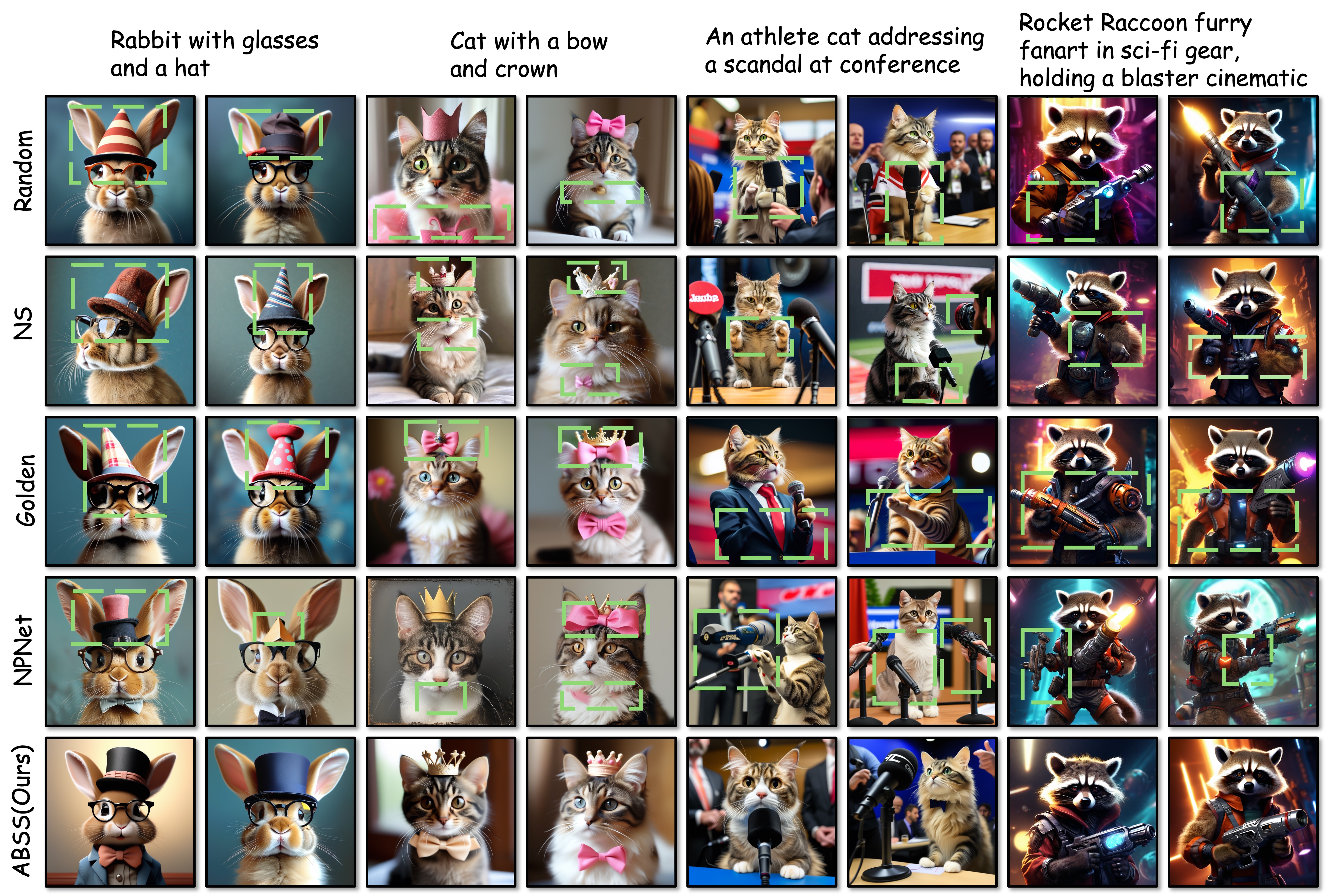}
    \vspace{-1mm}
    \caption{Qualitative comparisons on the Hunyuan-DiT backbone. Images in the same column are generated from the same prompt using different methods. Green dashed rectangles highlight the problematic generated regions. Results with other SD backbones are provided in Appendix~\ref{sec:results_q1_images}.}
    \label{fig:Q1}

    \vspace{2mm}

    \includegraphics[width=\textwidth]{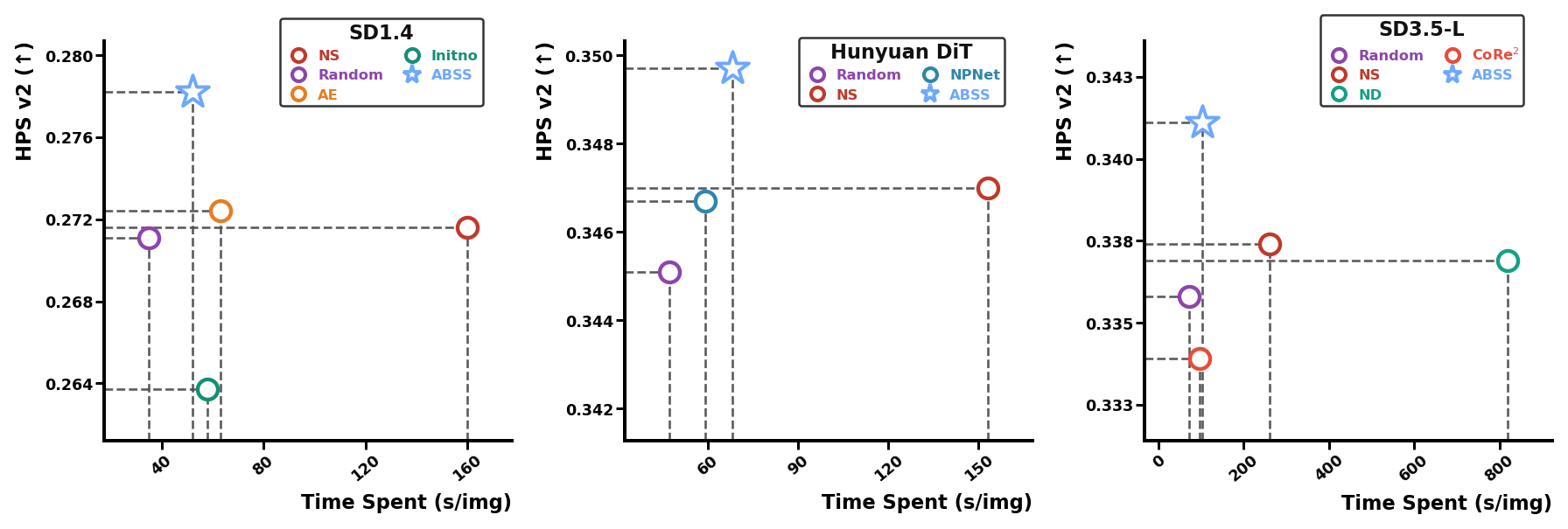}
    \vspace{-2mm}
    \caption{GPU latency comparison of average per-image generation on the \textit{InitNO} dataset across backbones. \textsc{ABSS} achieves the best quality-latency trade-off, obtaining the highest score with low latency. \textsc{Golden} is omitted because its overhead comes from validation-set seed search, while its per-image inference cost remains standard 50-step generation.}
    \label{fig:GPU}
    \vspace{-8mm}
\end{figure*}

For each main backbone, we include representative optimization or refinement baselines that are applicable to the corresponding model family, covering several directions beyond prompt-independent seed selection. On SD~1.4, we compare with Attend-and-Excite (\textsc{AE})~\citep{AEchef}, which refines attention maps during denoising for better text-image alignment, and \textsc{InitNO}~\citep{guo2024}, which optimizes the initial noise for each prompt. On SD~2.1, we include Noise Diffusion (\textsc{ND})~\citep{miao2024noisediffusion}, a noise-based optimization method for improving semantic faithfulness. For recent large-scale backbones, we compare with \textsc{CoRe}$^{2}$~\citep{shao2025core2} on SD~3.5 Large, which follows an iterative collect-reflect-refine framework, and Golden Noise (\textsc{NPNet})~\citep{zhou2024goldennoise} on Hunyuan-DiT, which learns to optimize diffusion noise. Detailed method settings and the corresponding NFE analysis are provided in Appendix  \ref{sec:supp_table1_setting}.

\begin{figure*}[!t]
    \centering\vspace{-6mm}
    \includegraphics[width=\linewidth, height=0.66\textheight, keepaspectratio]{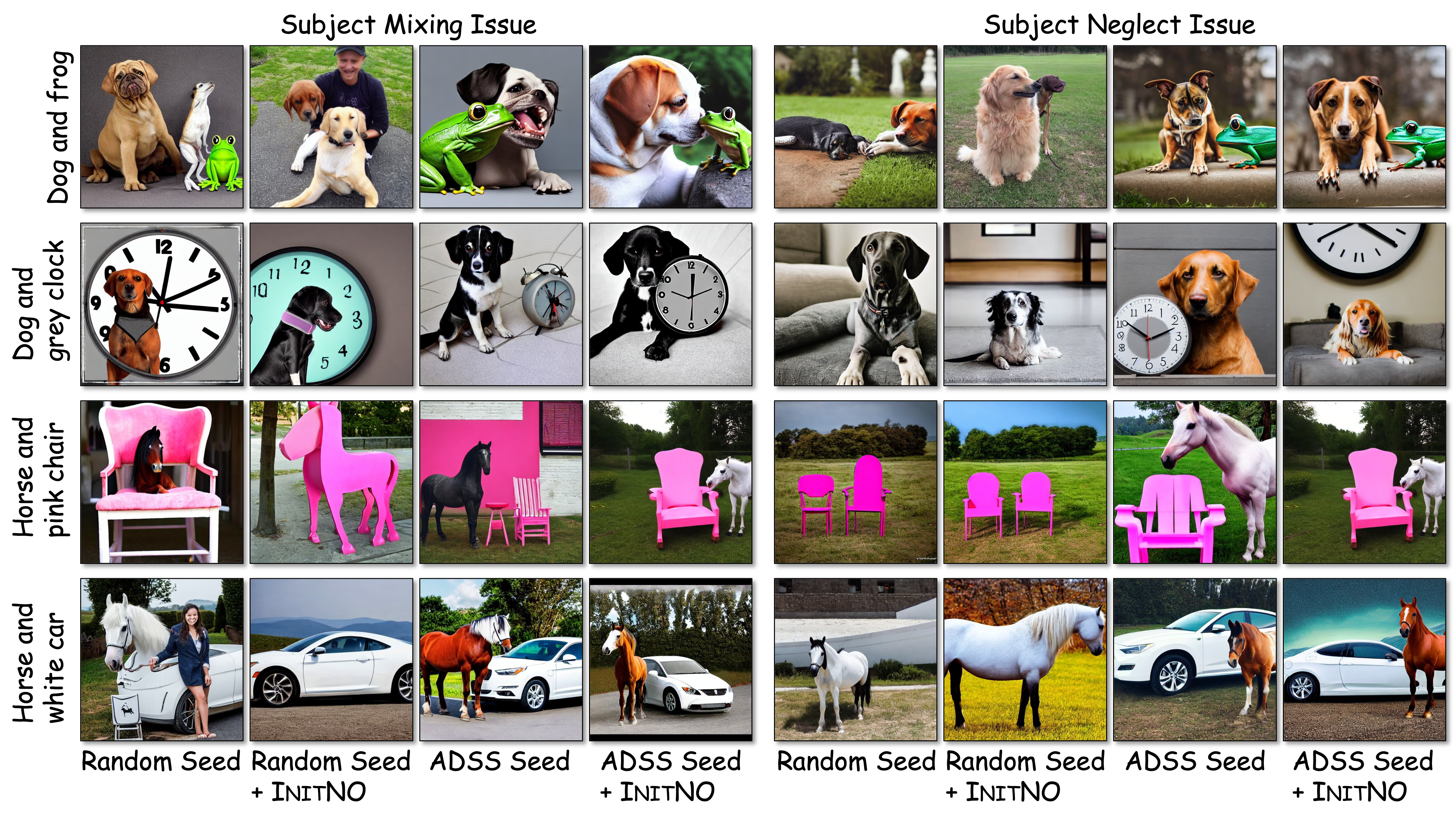}\vspace{-2mm}
    \caption{Comparison between \textsc{Random} and \textsc{ABSS} seeds for seed optimization by \textsc{InitNO} on challenging subject mixing and subject neglect issues across representative prompts.}
    \vspace{-6mm}
    \label{fig:add-on}
\end{figure*}

Compared with seed-selection baselines, \textsc{ABSS} provides a stronger quality-cost trade-off. \textsc{Random} is efficient but does not exploit prompt-dependent information, while \textsc{Golden} and \textsc{NS} require substantially larger search or validation overheads. In contrast, \textsc{ABSS} performs prompt-specific early screening and only continues the selected seeds to full denoising. As shown in Table~\ref{tab:diffusion_results_1}, \textsc{ABSS} consistently improves over \textsc{Random} across the main backbones and is competitive with or better than \textsc{Golden} and \textsc{NS} on most datasets and metrics. On SD~1.4, \textsc{ABSS} improves the \textit{InitNO} HPS from 0.2711 to 0.2782 and IR from 0.0612 to 0.1972 over \textsc{Random}. On Hunyuan-DiT, \textsc{ABSS} also achieves the best HPS and CLIP on all three datasets among the seed-selection methods. Importantly, this is achieved with only 71--73 NFE, compared with 333 NFE for \textsc{NS} and up to 209$^{\dagger}$ NFE for \textsc{Golden}, indicating that \textsc{ABSS} improves seed selection without introducing heavy search cost.

Compared with optimization and refinement baselines, \textsc{ABSS} achieves stronger overall performance with much lower inference overhead. On SD~1.4, \textsc{ABSS} improves over \textsc{AE} on the \textit{InitNO} from 0.2724 to 0.2782 in HPS and from 0.1277 to 0.1972 in IR, while using 73 NFE instead of 75$^{\dagger}$. On SD~2.1, although \textsc{ND} obtains slightly higher HPS and IR on \textit{InitNO}, it requires 550$^{\dagger}$ NFE; in contrast, \textsc{ABSS} uses only 73 NFE and achieves higher PickScore and CLIP across all three datasets under the same evaluation protocol.

\begin{wraptable}{h}{0.58\textwidth} 
  \centering\vspace{-4mm}
  \caption{Plug-in test with \textsc{ABSS}-selected seeds on \textit{InitNO} dataset.}
  \label{tab:plugin}\vspace{-2mm}
  \resizebox{0.58\textwidth}{!}{
  \begin{tabular}{l|c|cccc}
    \toprule
    Method & Version & HPS $\uparrow$ & IR $\uparrow$ & PickScore $\uparrow$ & CLIP $\uparrow$ \\
    \midrule
    \textsc{InitNO}          & 1.4  & 0.2637 & -0.2067 & 21.5729 & 0.2659 \\
    \enspace + \textsc{ABSS} & 1.4  & \textbf{0.2668} & \textbf{-0.1760} & \textbf{21.5803} & \textbf{0.2673} \\
    \midrule
    \textsc{InitNO}          & 1.5  & 0.2642 & -0.2057 & 21.5866 & 0.2651 \\
    \enspace + \textsc{ABSS} & 1.5  & \textbf{0.2680} & \textbf{-0.1535} & \textbf{21.6412} & \textbf{0.2684} \\
    \midrule
    \textsc{AE}              & 1.4  & 0.2724  & 0.1277 & 21.7440 & 0.2747 \\
    \enspace + \textsc{ABSS} & 1.4  & \textbf{0.2803}  & \textbf{0.3148} & \textbf{21.8807} & \textbf{0.2784} \\
    \midrule
    \textsc{AE}              & 1.5  & 0.2731 & 0.1494 & 21.7586 & 0.2750 \\
    \enspace + \textsc{ABSS} & 1.5  & \textbf{0.2818} & \textbf{0.3250} & \textbf{21.8958} & \textbf{0.2789} \\
    \bottomrule
  \end{tabular}}\vspace{-2mm}
\end{wraptable}

Our advantage is more pronounced on recent backbones. On Hunyuan-DiT, \textsc{ABSS} surpasses \textsc{NPNet}: it improves \textit{DrawBench} by +0.0047 HPS, +0.1045 IR, and +0.0064 CLIP, and improves \textit{InitNO} PickScore by +0.0768. On SD~3.5 Large, \textsc{ABSS} outperforms \textsc{CoRe}$^{2}$ on most metrics, improving \textit{DrawBench} HPS by +0.0059, \textit{InitNO} HPS/IR by +0.0072/+0.0256, and \textit{Pick-a-Pic} HPS/IR/PickScore by +0.0062/+0.0790/+0.2066. Although \textsc{ND} can be better on a few SD~3.5 Large metrics, it requires 550$^{\dagger}$ NFE compared with 71 NFE for \textsc{ABSS}. Importantly, \textsc{ABSS} achieves these wins with strong performance and low inference cost. Beyond the NFE comparison in Table~\ref{tab:diffusion_results_1}, Figure~\ref{fig:GPU} shows this gap in wall-clock latency: \textsc{ND} is almost an order of magnitude slower than \textsc{ABSS}, whereas \textsc{NPNet} and \textsc{CoRe}$^{2}$ are slightly faster but yield weaker quality. These results demonstrate that \textsc{ABSS} provides a stronger quality--efficiency trade-off. Additional significance tests and NDCG results are provided in Appendix~\ref{sec:supp_q1}.

\subsection{Plug-in to Seed/Initial-Noise Optimization (Q2)}
\looseness-1Here we test whether \textsc{ABSS} can be combined with state-of-the-art inference-time optimization to provide an extra boost. We consider two recent methods as baselines. \textsc{InitNO}~\citep{guo2024} explicitly optimizes the initial noise per prompt under a chosen objective and then samples from the optimized initialization. \textsc{Attend-and-Excite} (AE)~\citep{AEchef} is an inference-time attention-based guidance method that steers cross-attention toward subject tokens to improve prompt faithfulness without retraining. In our integration, \textsc{ABSS} serves only as a seed filter; baseline pipelines keep their default losses, schedulers, and hyperparameters—we simply replace uniform random seeds with \textsc{ABSS}-selected seeds. 

Table~\ref{tab:plugin} shows plugging \textsc{ABSS} into \textsc{InitNO} and \textsc{AE} yields consistent across-the-board improvements in all four metrics, demonstrating strong compatibility with state-of-the-art latent optimization methods.
For \textsc{InitNO} on SD~1.4/1.5, IR improves from -0.2067/-0.2057 to -0.1760/-0.1535 and PickScore increases from 21.5729/21.5866 to 21.5803/21.6412, with HPS/CLIP also increased. For \textsc{AE} on SD~1.4/1.5, the gains are larger: PickScore rises from 21.7440/21.7586 to 21.8807/21.8958 and IR from 0.1277/0.1494 to 0.3148/0.3250, accompanied by increases in HPS and CLIP. These results indicate that \textsc{ABSS} supplies better starting latents and particularly strengthens inference-time attention guidance.

\begin{wrapfigure}[23]{r}{0.5\textwidth}
  \vspace{-6mm}
  \centering
  \includegraphics[width=\linewidth]{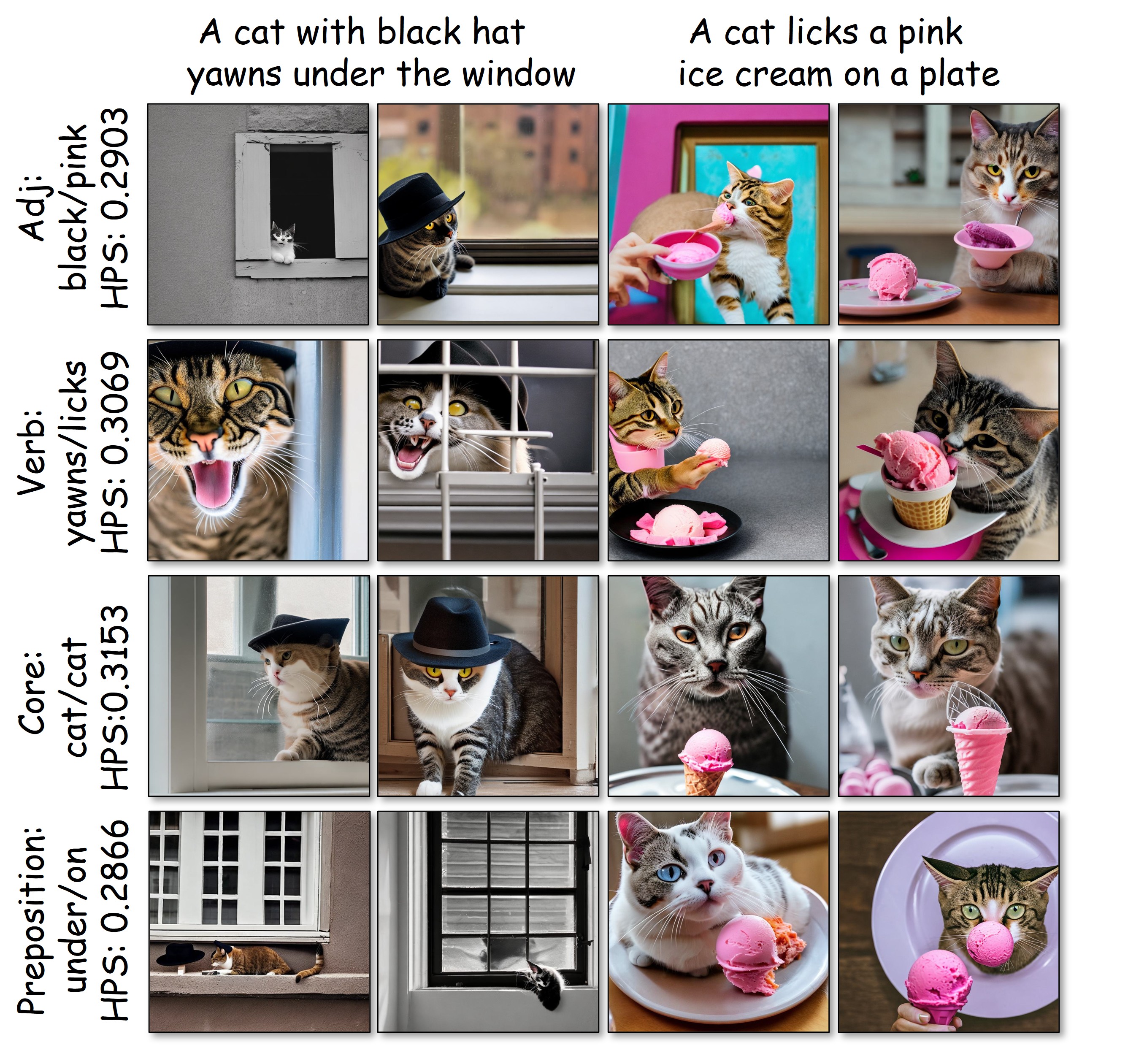}
  \caption{Generated images with different token-type focuses. Focusing on the core tokens produces the most semantically faithful and visually plausible images, achieving the highest HPS score.}
  \label{fig:q3}
    \vspace{-6mm}
\end{wrapfigure}
In Figure~\ref{fig:add-on}, we show that, although \textsc{InitNO} is designed to alleviate subject neglect (the primary subject does not appear) and subject mixing (two or more objects are conflated), certain seeds still exhibit these failures even after optimization. For instance, in the horse--pink-chair case: without optimization the horse and chair are entangled; after optimization a pink ``chair--horse'' emerges; and in the missing-subject case the horse remains absent despite optimization. Our \textsc{ABSS} mitigates these failures by filtering out such bad seeds, thereby addressing these issues and strengthening \textsc{InitNO} or other pipelines.

\subsection{Ablation on Token Type (Q3)}
Beyond core tokens, we ablate alternative token types to answer Q3 and test whether core-token scoring is sufficient. For the prompt ``\textit{A cat with black hat yawns under the window},'' we label \emph{black} as an adjective, \emph{cat} as a core token, \emph{yawns} as a verb, and \emph{under} as a preposition. On \textit{DrawBench} (rich in adjectives and verbs), we find that scoring core tokens yields the best overall performance, improving HPS and IR over alternative token types. More detailed ablation settings and results are provided in Appendix~\ref{sec:supp_q3}. We also visualize \textsc{ABSS} with different token type focuses. Figure~\ref{fig:q3} shows that emphasizing \emph{verbs} (e.g., ``yawns,'' licks'') promotes motion cues but often harms object details, yielding incomplete subjects and lower HPS. Emphasizing \emph{adjectives} (e.g., ``black,'' pink'') spreads color/style broadly—often bleeding into background—while the main subject becomes less faithful. Focusing on \emph{preposition} improves relative placement (e.g., window/plate context) but increases missing or fragmented foreground objects. In contrast, focusing on \emph{core tokens} best preserves subject integrity and semantics, achieving the strongest HPS; thus, \textsc{ABSS} only needs early attention on core tokens.

\section{Conclusion}
We revisited the seed effect in T2I and showed that early cross-attention to core tokens strongly predicts prompt alignment and image quality. Building on this, we proposed \textsc{ABSS}, a training-free, plug-and-play procedure that globally ranks seeds via a core-token concentration score and works across different Diffusion/Flow based variants. Experiments on three benchmarks demonstrated consistent gains in faithfulness and visual quality, improving human preference and alignment metrics with minimal overhead. \textsc{ABSS} also serves as a lightweight preselection layer that complements seed-optimization and rollback/refinement pipelines. These results positioned attention-based seed selection as a practical primitive for robust, cost-efficient T2I at scale.

\section{Acknowledgment}
We gratefully acknowledge computational support from the Brandeis High Performance Computing Cluster (HPCC), which is partially supported by NSF OAC-1920147. We also thank the anonymous reviewers and the area chair for their valuable feedback and constructive suggestions.

\section{Limitation}
Although \textsc{ABSS} is training-free and broadly applicable across different diffusion backbones, it relies on access to early attention maps or intermediate attention statistics. Therefore, applying \textsc{ABSS} to closed-source systems or architectures where such signals are inaccessible may require additional engineering. In addition, \textsc{ABSS} assumes that the core subject tokens in a prompt can be reasonably identified. Although our robustness experiments show that \textsc{ABSS} remains effective under noisy core-token annotations, its seed-ranking signal may become less reliable for highly ambiguous prompts or prompts involving many fine-grained subject relations.

\bibliographystyle{plainnat}  
\bibliography{references}

\clearpage
\appendix

\section*{Appendix}

\section{Definition of Core Tokens}
\label{app:core_tokens}
In \textsc{ABSS}, core tokens are the content-bearing words that specify the primary semantic entities in a prompt, i.e., the main objects or subjects that should dominate the image. For small prompt sets, users can directly specify them by selecting the main subject/object words (typically nouns or short noun phrases). For larger prompt collections, we automatically identify core tokens using a large language model (LLM) as a one-time preprocessing step.

\paragraph{LLM-based core-token identification at scale.}
For large prompt collections, we identify core tokens automatically using an LLM applied once per prompt. Concretely, we use GPT-5.1 with a simple instruction to extract up to three core words that denote the main objects or subjects expected to dominate the image, while labeling other content words into coarse categories (adjectives, verbs, or prepositions). This procedure is fully automatic and incurs negligible overhead compared to diffusion sampling, since it is run only once per prompt. Moreover, our results show that \textsc{ABSS} is robust to moderate noise in this step: even approximate core-token identification already yields consistent improvements, suggesting that the method does not rely on perfectly accurate annotation. In our implementation, processing 100 prompts with the LLM takes slightly over one minute and produces highly accurate outputs under manual inspection, further indicating that the preprocessing is both cheap and reliable.

\paragraph{Prompt template used for LLM extraction.}

We use the following prompt template for each text prompt:
\begin{mdframed}
\small
\textbf{Instruction.} Assuming you are an expert in text-to-image prompting and visual semantics, given a text-to-image prompt, identify up to three core words that denote the main objects or subjects that would occupy most of the image. Output these as \texttt{core\_tokens}. Then label the remaining content words into \texttt{adjectives}, \texttt{verbs}, and \texttt{prepositions} (ignore articles and stopwords).\\[4pt]

\textbf{Prompt 1.} \texttt{Kneeling cat knight portrait finely detailed armor intricate design silver silk cinematic lighting}\\
\textbf{Output 1 (JSON).}
\begin{verbatim}
{"core_tokens": ["cat", "knight", "armor"],
 "adjectives": ["kneeling", "portrait", "finely", "detailed", "intricate",
                "silver", "silk", "cinematic"],
 "verbs": [],
 "prepositions": []}
\end{verbatim}
\end{mdframed}
\begin{mdframed}
\textbf{Prompt 2.} \texttt{A galactic eldritch squid towering over the planet Earth stars galaxies and nebulas in the background photorealistic 8k}\\
\textbf{Output 2 (JSON).}
\begin{verbatim}
{"core_tokens": ["squid", "Earth", "nebulas"],
 "adjectives": ["galactic", "eldritch", "towering", "planet",
                "photorealistic", "8k"],
 "verbs": [],
 "prepositions": ["over", "in"]}
\end{verbatim}
\end{mdframed}
We then map the extracted core words to tokenizer indices using the underlying text encoder tokenizer (e.g., CLIP BPE), and compute core-token cross-attention statistics accordingly.
\paragraph{Robustness to Noisy Core-Token Annotations.}
\label{app:core_token_robustness}
We further evaluate the robustness of \textsc{ABSS} to noisy core-token annotations. Since \textsc{ABSS} only relies on a small set of main subject tokens, it is expected to be relatively insensitive to moderate annotation errors. To verify this, we intentionally corrupt 50\% of the core-token annotations and evaluate \textsc{ABSS} on FLUX.1 using the \textit{InitNO} dataset. As shown in Table~\ref{tab:core_token_robustness}, \textsc{ABSS} still improves over random seed selection across all metrics even when half of the core-token annotations are corrupted, suggesting that its effectiveness does not rely on perfectly annotated core tokens.

\begin{table}[!t]
\centering
\caption{Robustness of \textsc{ABSS} to noisy core-token annotations on FLUX.1 using the \textit{InitNO} dataset. We intentionally corrupt 50\% of the core-token annotations. Even under this noisy setting, \textsc{ABSS} still improves over random seed selection across all metrics.}
\label{tab:core_token_robustness}
\setlength{\tabcolsep}{6pt}
\begin{tabular}{lcccc}
\toprule
\textbf{Setting} & \textbf{HPS} $\uparrow$ & \textbf{IR} $\uparrow$ & \textbf{PickScore} $\uparrow$ & \textbf{CLIP} $\uparrow$ \\
\midrule
\textsc{Random} & 0.3364 & 1.5945 & 23.8038 & 0.2967 \\
\textsc{ABSS} w/ 50\% noisy tokens & \textbf{0.3382} & \textbf{1.5958} & \textbf{23.8134} & \textbf{0.2974} \\
\bottomrule
\end{tabular}
\end{table}

\section{Extend ABSS to DiT based architecture}
\label{app:dit_def}
Diffusion Transformers (DiTs) have recently emerged as a strong alternative to UNet-based diffusion backbones. In particular, DiT~\citep{peebles2022dit} demonstrates that scaling transformer capacity and training can yield highly competitive---and often state-of-the-art---generation quality, making transformer-based diffusion architectures increasingly popular in modern text-to-image systems. Given this trend, it is important to understand whether our attention-based seed scoring principle is specific to UNets, or whether it generalizes to DiT-style backbones.

In the main paper, we show that ABSS is effective on UNet-based Stable Diffusion across both the 1.x and 2.x series, where cross-attention directly reveals how image latents focus on core text tokens in early denoising. We further show that the same principle extends to DiT-style backbones: although DiT lacks
UNet cross-attention blocks, its in-context conditioning jointly processes image and text tokens via self-attention, enabling an analogous image-to-core-text signal for seed scoring. Specifically, we investigate \textsc{ABSS} on SD~3.0 and FLUX to assess robustness across modern architectures and training objectives. Although FLUX and SD~3.0 is trained with a flow-matching objective, \textsc{ABSS} only relies on early-step attention under text conditioning, making it largely architecture-driven rather than objective-specific. Below we formalize the ABSS score under DiT conditioning.

\paragraph{Aggregated Self-Attention Map.}
Unlike UNet-based Stable Diffusion, DiT patchifies the latent/image into $M$ patch tokens and concatenates them with the $N$ text tokens, then applies standard self-attention over the joint sequence. Although DiT has no explicit cross-attention blocks, its self-attention matrix still contains a well-defined image-to-text attention signal.

Concretely, DiT represents the current latent state at timestep $t$ as the combination fo $M$ patch image tokens $x_{t}^{(l-1)}\in\mathbb{R}^{M\times d}$ and the conditional prompt as $N$ text tokens $c\in\mathbb{R}^{N\times d}$. These tokens are concatenated into a single sequence
\[
X_{t}^{(l-1)}=
\begin{bmatrix}
x_{t}^{(l-1)}\\
c
\end{bmatrix}
\in\mathbb{R}^{(M+N)\times d},
\]
where indices $1,\ldots,M$ correspond to image patch tokens and $M+1,\ldots,M+N$ correspond to text tokens.
The self-attention matrix at timestep $t$ and layer $(l)$ is
\[
A_{t}^{(l)}=\mathrm{softmax}\!\left(\frac{Q_{t}^{(l)}(K_{t}^{(l)})^\top}{\sqrt{d_k}}\right)
\in\mathbb{R}^{(M+N)\times(M+N)},
\]
where $Q_{t}^{(l)},K_{t}^{(l)}$ are the query/key projections of $X_{t}^{(l-1)}$ and $A_{t}^{(l)}[u,v]$ is the
attention weight from token $u$ to token $v$. In particular, $A_{t}^{(l)}[i,\,M+j]$ with
$i\in\{1,\ldots,M\}$ and $j\in\{1,\ldots,N\}$ measures how strongly the $i$-th image patch token attends to
the $j$-th text token, analogous to the token distribution at a spatial location $(h,w)$ in the UNet case.

Instead of stacking self-attention maps across multiple layers, we empirically find that hooking a single intermediate transformer layer $l^\star$ yields the best ABSS performance. Given a seed $s$ and timestep $t$,
we denote the hooked self-attention matrix by $\tilde{A}_t^{s}\triangleq \tilde{A}_{t}^{s,(l^\star)}$ and use its image-to-text
entries $\tilde{A}_t^{s}[i,\,M+j]$ as the attention signal for seed scoring.

\paragraph{Core-token sorting: averaging, smoothing, and pooling.}
Given the hooked self-attention matrix $\tilde{A}^{\,s}_{t}\in\mathbb{R}^{(M+N)\times(M+N)}$ for seed $s$ at timestep $t$,
we first average over image tokens to obtain a scalar score for each text token $i\in\{1,\ldots,N\}$:
\[
M^{s}_{t}[i]
\;=\;
\frac{1}{M}\sum_{j=1}^{M}\tilde{A}^{\,s}_{t}[j,\,M+i].
\]
We then smooth this token-level score vector along the text-token axis using a normalized Gaussian kernel:
\[
\widehat{M}^{s}_{t} \;=\; G_{\sigma}\ast_{\mathrm{refl}}\, M^{s}_{t},
\]
where $G_{\sigma}$ is a normalized 1D Gaussian kernel and $\ast_{\mathrm{refl}}$ denotes discrete convolution with reflection padding.
Finally, given the core-token index set $B\subseteq\{1,\ldots,N\}$, we define the core-token concentration score as
\[
M_{t}^{s}(B)
\;=\;
\frac{1}{|B|}\sum_{i\in B}\widehat{M}^{s}_{t}[i],
\]
which we use in ABSS to pre-screen and rank candidate seeds for DiT-based models.

\section{Setup, Supplementary Experiments and Analyses}
\label{sec:supp}

\subsection{Detailed Generation and Evaluation Settings for Table~\ref{tab:diffusion_results_1}}
\label{sec:supp_table1_setting}

\paragraph{ABSS Attention extraction and sampling settings.}
We provide additional implementation details for reproducing Table~\hyperref[tab:diffusion_results_1]{1}. 
For SD~1.4/1.5, we aggregate cross-attention maps across all layers and heads at a spatial resolution of $16{\times}16$, while for SD~2.0/2.1 we use $24{\times}24$. 
For transformer-based backbones, we extract attention from a single representative transformer block: the 12th block for FLUX.1, SD~3.0, and Hunyuan-DiT, and the 18th block for SD~3.5-Large. Unless otherwise specified, we use classifier-free guidance with guidance scale $7.5$ and 50 inference steps for all backbones. Image resolution is fixed per model series: SD~1.x uses $512{\times}512$, SD~2.x uses $768{\times}768$, and FLUX.1, SD~3.0, SD~3.5-Large, and Hunyuan-DiT use $1024{\times}1024$.

\paragraph{Comparison method settings and averaging protocol.}
We summarize the method-specific settings for comparison methods as follows. 
For a fair comparison, all shared generation configurations, including image resolution, classifier-free guidance scale, and the number of inference steps, are aligned with ABSS. 
We also report a coarse NFE estimate for each method, where one full denoising-model forward pass is counted as one NFE.
\begin{enumerate}[leftmargin=*]
    \item \textsc{Golden}. 
    
    We use the first 100 prompts from \textit{InitNO} and \textit{DrawBench}, and the first 50 prompts from \textit{Pick-a-Pic} as a validation set to extract the ``golden seeds''; all remaining prompts are used for evaluation. Specifically, \textsc{Golden} ranks candidate seeds by their average HPS-v2 score on the validation set, and then applies the top-ranked seeds to all test prompts in the same dataset. The validation search cost is amortized over the reported test generations. 
    For example, on \textit{DrawBench},
    \[
    \text{NFE} = 50 + \frac{100 \times 10 \times 50}{100 \times 3} \approx 216.7\dagger.
    \]
    Averaged across datasets, we report the coarse NFE of \textsc{Golden} as $209^{\dagger}$, where $\dagger$ indicates that the method involves additional HPS-v2 evaluation overhead during validation-set seed ranking.
    \item \textsc{NS}. 
    For a fair comparison with the \textsc{ABSS} setting using a 10-seed pool, we use $K{=}10$ noise candidates per prompt instead of the default candidate number of 100. 
    Following the official implementation, each candidate is ranked by its DDIM inversion stability: we first run 50-step DDIM sampling from $z_T$ to $z_0$, then perform 50-step DDIM inversion back to $z_T'$, and compute $\cos(z_T,z_T')$ as the ranking score. 
    We then select the top-3 candidates and directly report their already generated images. 
    Thus, the coarse NFE per reported image is
    \[
    \text{NFE} = \frac{10 \times (50 + 50)}{3} \approx 333.3.
    \]

    \item \textsc{InitNO}. 
    We follow the official \textsc{InitNO} implementation, optimizing the initial noise mean and log-variance with Adam using learning rate $1{\times}10^{-2}$. 
    We use the default attention thresholds $\tau_{\text{cross}}{=}0.2$ and $\tau_{\text{self}}{=}0.3$, with at most 5 restart rounds of 10 optimization steps. 
    According to the pseudo-code in \textsc{InitNO}, each denoising optimization step invokes one full denoising-model forward pass. 
    Thus, the maximum optimization cost is $5 \times 10$ NFEs, and the final image generation uses another 50 denoising steps. 
    Since the threshold-based stopping criterion may terminate the optimization early, the actual latency can be lower in practice. 
    For simplicity, we report the coarse NFE as
    \[
    \text{NFE} = 5 \times 10 + 50 = 100^{\dagger},
    \]
    where $\dagger$ indicates additional latent/noise optimization overhead beyond standard denoising.
    
    \item \textsc{AE}. 
    We follow the official \textsc{AE} latent optimization setting, using attention maps at $16{\times}16$ resolution. 
    Latent updates are applied within the first 25 denoising steps with scale factor 20, and iterative refinement is triggered at steps 10 and 20 using thresholds $\tau_{\text{cross}}{=}0.2$ and $\tau_{\text{self}}{=}0.3$, with at most 20 refinement steps. 
    Final images are decoded after the standard 50-step denoising process. 
    Since \textsc{AE} performs one additional attention/loss forward pass for each of the first 25 denoising steps, the coarse NFE is computed as
    \[
    \text{NFE} = 50 + 25 = 75^{\dagger}.
    \]
    For simplicity, we do not include the threshold-triggered iterative refinement steps in this coarse NFE estimate, as they are only invoked when the attention loss does not satisfy the specified thresholds. 
    Here, $\dagger$ indicates additional latent-optimization overhead beyond standard denoising.

    \item \textsc{ND}. 
    We do not follow the default \textsc{ND} setting, as its original configuration uses 50 optimization epochs and 50 noise candidates, which incurs very high computational cost. Instead, for a fairer comparison, we reduce the setting to 10 optimization epochs and 10 noise candidates. 
    We otherwise follow the official \textsc{ND} implementation, which uses VQAScore with \texttt{clip-flant5-xl} as the outer objective. 
    At each optimization epoch, \textsc{ND} updates the noise based on the selected candidate, and the final image is generated with the same 50-step denoising setting as \textsc{ABSS}. 
    The coarse NFE is therefore computed as
    \[
    \text{NFE} = 10 \times 50 + 50 = 550^{\dagger},
    \]
    where the first term denotes the 10 optimization epochs and the second term denotes the final 50-step image generation. 
    Here, $\dagger$ indicates additional overhead from gradient-cache updates and VQAScore evaluation, which is not fully captured by the coarse NFE.

    \item \textsc{CoRe$^2$}. 
    We follow the official \textsc{CoRe$^2$} implementation on SD~3.5-Large with the released noise model checkpoint. 
    The refinement module uses a LoRA-based PromptSD35Net with 28 LoRA slots and rank 64. We set the weak-to-strong guidance scale to 1.5 and apply the refinement branch at every denoising step. For each prompt, we sample 3 images with different random seeds. Although the refinement branch is applied during inference, it does not introduce an additional full denoising-model forward pass through all blocks. Therefore, the number of full denoising steps remains 50, and we report its coarse NFE as $50^{\dagger *}$, where $\dagger$ means the refine step within each denoising step and $*$ indicates additional training cost for the Collect-and-Reflect refinement module.

    \item \textsc{NPNet}. 
    We follow the official \textsc{NPNet} inference pipeline and use the released pretrained noise-prompt model for each backbone. 
    For Hunyuan-DiT, we use the DiT branch with the released \texttt{dit.pth} checkpoint to predict the golden initial noise. 
    For each prompt, we generate 3 golden-noise samples with deterministic seeds and use the generated images as the \textsc{NPNet} baseline. 
    During inference, \textsc{NPNet} only applies a lightweight auxiliary module once at the beginning to predict or adjust the initial noise, while the subsequent denoising process still follows the standard 50-step generation. 
    Since this auxiliary module does not require an additional full denoising-model forward pass, we report its coarse NFE as $50^{\dagger *}$, where $*$ indicates the use of a separately trained auxiliary module whose training cost is not counted in NFE and $\dagger$ represents the extra initial NPNet noise optimization during inference.
    \item \textsc{ABSS}. 
    We use a seed pool of 10 candidates per prompt and select the top-3 seeds for final image generation. 
    For SD 1.x and SD 2.x, attention maps are collected at the 10th denoising step across all layers and heads, using spatial resolution $16{\times}16$ for SD 1.x and $24{\times}24$ for SD 2.0/2.1. 
    Since this requires a full forward pass, the coarse NFE per reported image is
    \[
    \text{NFE} = \frac{10 \times 10 + 3 \times 40}{3} \approx 73.3.
    \]
    For FLUX.1, Hunyuan-DiT, SD 3.0, and SD 3.5-Large, we use truncated forward screening by extracting attention from an intermediate block. 
    For example, with attention collected at the 12th block out of 30 total blocks, the coarse NFE becomes
    \[
    \text{NFE} = \frac{10 \times (9 + 12/30) + 3 \times 40}{3} \approx 71.3.
    \]
    Accordingly, we report the coarse NFE of \textsc{ABSS} as approximately 73 for the U-Net backbones and 71 for the DiT-based backbones.
\end{enumerate}
\begin{table*}[!t]
\centering
\caption{Additional results on the remaining backbones. We report mean top-$k$ performance with $k=3$ from a 10-seed pool; the best result among \textsc{Random}, \textsc{Golden}, and \textsc{ABSS} is bolded for each metric.}
\label{tab:diffusion_results}
\setlength{\tabcolsep}{4pt}
\resizebox{\textwidth}{!}{%
\begin{tabular}{cl|cccc|cccc|cccc}
\toprule
\multicolumn{2}{c|}{} &
\multicolumn{4}{c|}{\textit{DrawBench}} &
\multicolumn{4}{c|}{\textit{InitNO}} &
\multicolumn{4}{c}{\textit{Pick-a-Pic}} \\
\cmidrule(lr){3-6}\cmidrule(lr){7-10}\cmidrule(lr){11-14}
\textbf{Version} & \textbf{Method} &
HPS $\uparrow$ & IR $\uparrow$ & PickScore $\uparrow$ & CLIP $\uparrow$ &
HPS $\uparrow$ & IR $\uparrow$ & PickScore $\uparrow$ & CLIP $\uparrow$ &
HPS $\uparrow$ & IR $\uparrow$ & PickScore $\uparrow$ & CLIP $\uparrow$ \\
\midrule

\multirow{3}{*}{1.5}
  & \textsc{Random} & 0.2428  & -0.2651 & 20.5737 & 0.2569  &
             0.2702 & -0.0286 & 21.7658 & 0.2712 & 0.2446  & \textbf{-0.1732} & 20.2062 & 0.2541 \\
  & \textsc{Golden} & \textbf{0.2488} & -0.1678 & \textbf{20.6277} & 0.2558 &
             0.2754 & 0.0189 & 21.8361 & 0.2723 & 0.2461 & -0.2261 & \textbf{20.2213} & 0.2546 \\
  & \textsc{ABSS}   & 0.2484 & \textbf{-0.1666} & 20.6194 & \textbf{0.2576} &
             \textbf{0.2757} & \textbf{0.1191} & \textbf{21.8381} & \textbf{0.2735} & \textbf{0.2465} & -0.2119 & 20.2203 & \textbf{0.2551} \\
\midrule

\multirow{3}{*}{2.0}
  & \textsc{Random} & 0.2504 & 0.0214 & 20.9348 & 0.2662 &
             0.2814 & 0.8255 & 22.1335 & 0.2926 & 0.2612  & 0.1038 & 20.5830 & 0.2689 \\
  & \textsc{Golden} & 0.2601 & \textbf{0.2014} & \textbf{21.0551} & 0.2668 &
             0.2936 & 1.0140 & 22.3182 & 0.2942 & \textbf{0.2705} & \textbf{0.2732} &  \textbf{20.7832}& 0.2703 \\
  & \textsc{ABSS}   & \textbf{0.2604} & 0.1558 & 21.0299 & \textbf{0.2669} &
             \textbf{0.2942} & \textbf{1.0489} & \textbf{22.3478} & \textbf{0.2955} & 0.2702 & 0.2578 & 20.7765 & \textbf{0.2714} \\
\midrule

\multirow{3}{*}{3.0}
  & \textsc{Random} & 0.2963  & 0.9782 & 22.0641 & 0.2862 &
             0.3405 & 1.8090 & 23.6124  & 0.3052  &
             0.3110 & 1.0498 & 22.0072 & 0.2750 \\
  & \textsc{Golden} & 0.2987 & 1.0203 & \textbf{22.1714} & 0.2878 &
             0.3422 & 1.8089 & 23.6061 & \textbf{0.3065} &
             0.3129 & 1.0630 & 22.0813 & \textbf{0.2779} \\
  & \textsc{ABSS}   & \textbf{0.2991} & \textbf{1.0292} & 22.1453 & \textbf{0.2881} &
            \textbf{0.3440} & \textbf{1.8091} & \textbf{23.6398} & 0.3057 &
             \textbf{0.3144} & \textbf{1.0757} & \textbf{22.0831} & 0.2753 \\
\midrule

\multirow{3}{*}{FLUX}
  & \textsc{Random} & 0.2971  & 0.7093 & 22.0118  & 0.2624 &
             0.3364 & 1.5945 & 23.8038 & 0.2967 &
             0.3094 & 0.7997 & 22.0647 & 0.2568 \\
  & \textsc{Golden} & 0.2986 & 0.6618 & 21.9796 & 0.2616 &
             0.3382 & 1.6125 & 23.8249 & 0.2960 &
             0.3088 & \textbf{0.8161} & 22.0442 & 0.2572 \\
  & \textsc{ABSS}   & \textbf{0.2996} & \textbf{0.7273}  & \textbf{22.0242} & \textbf{0.2641} &
             \textbf{0.3408} & \textbf{1.6517} & \textbf{23.8505} & \textbf{0.2980} &
             \textbf{0.3114} & 0.8060 & \textbf{22.0716} & \textbf{0.2581} \\
\bottomrule
\end{tabular}\vspace{-6mm}
}
\end{table*}

\subsection{Additional Results}
\label{sec:results_q1_table}

We additionally provide results on the remaining backbones, including SD~1.5, SD~2.0, SD~3.0 Medium, and FLUX.1, to complement the main-paper comparisons on SD~1.4, SD~2.1, SD~3.5 Large, and Hunyuan-DiT. As shown in Table~\ref{tab:diffusion_results}, \textsc{ABSS} generally improves over \textsc{Random} and remains competitive with or better than \textsc{Golden} across most datasets and metrics. On SD~1.5, \textsc{ABSS} improves IR over \textsc{Random} on \textit{DrawBench} and \textit{InitNO}, from $-0.2651$ to $-0.1666$ and from $-0.0286$ to $0.1191$, respectively. On SD~2.0, \textsc{ABSS} achieves the best HPS and CLIP on \textit{DrawBench}, and further improves \textit{InitNO} performance, increasing IR from $0.8255$ to $1.0489$ over \textsc{Random}. 

The gains are also consistent on stronger backbones. On SD~3.0 Medium, \textsc{ABSS} obtains the best HPS and IR on \textit{DrawBench}, improving IR from $0.9782$ to $1.0292$, and achieves the best HPS, IR, and PickScore on \textit{Pick-a-Pic}. The improvement is particularly strong on FLUX.1, where \textsc{ABSS} outperforms both \textsc{Random} and \textsc{Golden} on 11 out of 12 metrics. For example, on \textit{DrawBench}, \textsc{ABSS} improves HPS from $0.2971$ to $0.2996$, IR from $0.7093$ to $0.7273$, PickScore from $22.0118$ to $22.0242$, and CLIP from $0.2624$ to $0.2641$ over \textsc{Random}. These additional results further support that \textsc{ABSS} generalizes across different Stable Diffusion versions and recent non-U-Net backbones.

\begin{figure}[!t]
  \centering
  \vspace{-2mm}
  \includegraphics[width=0.8\textwidth]{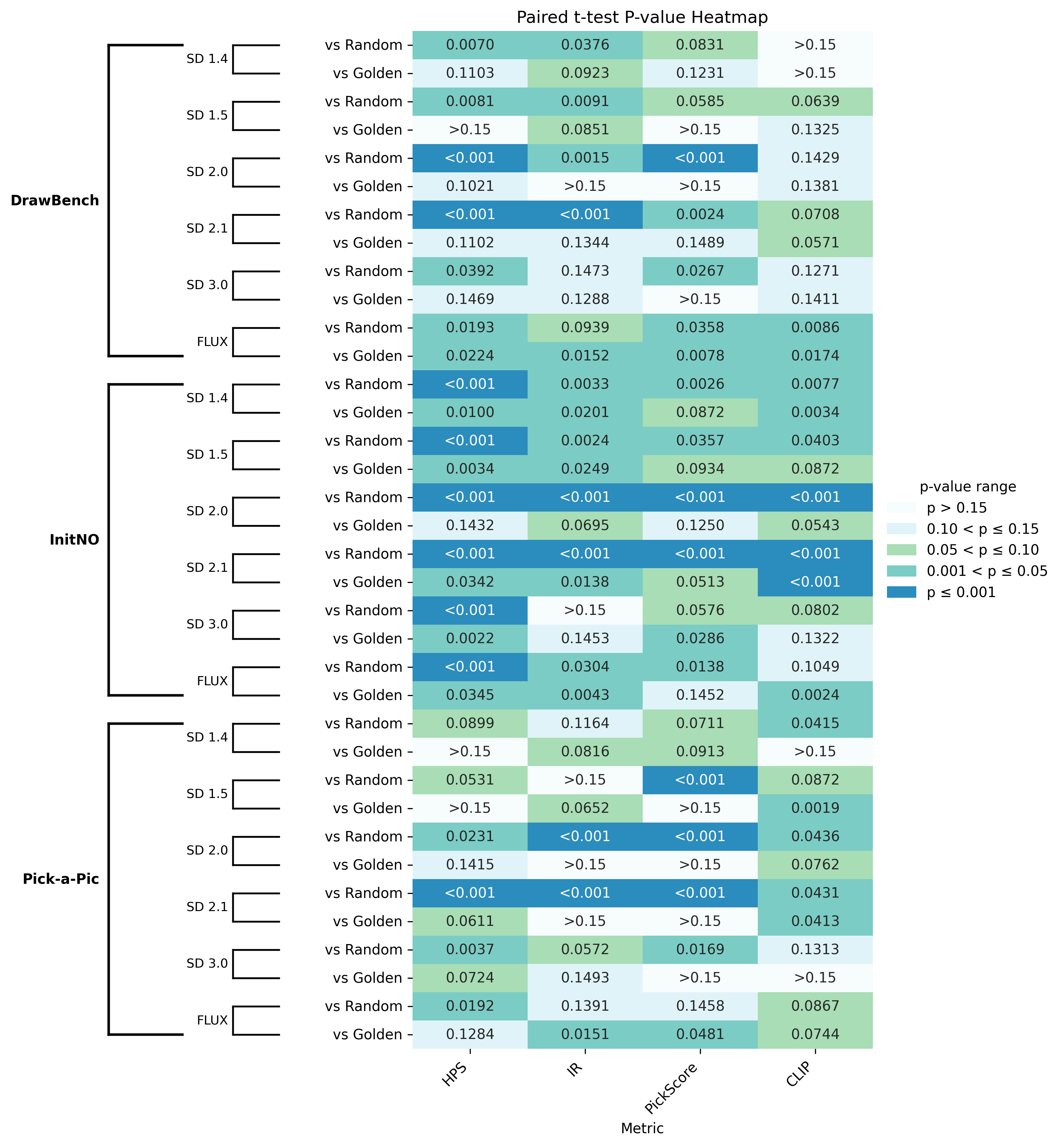}
  \vspace{-2mm}
  \caption{Paired t-test $p$-values for the entries in Table~\ref{tab:diffusion_results} (computed across prompts).}
  \label{fig:t1_ptest}
  \vspace{-3mm}
\end{figure}

\subsection{Additional Evidence for Baseline Comparisons for Q1}
\label{sec:supp_q1}

To assess the stability of our conclusions, we provide additional statistical and diagnostic evidence for \textsc{ABSS} beyond the main comparisons. Specifically, we validate \textsc{ABSS} from three complementary perspectives: (i) paired significance testing on the overall metrics, (ii) ranking agreement with the HPS-induced seed ordering, and (iii) sensitivity to the evaluation timestep $t$.

\paragraph{Paired t-test on Table~1.}

We further conduct paired t-tests on the entries in Table~\ref{tab:diffusion_results} and parts of entries in Table~\ref{tab:diffusion_results_1} across prompts. For each backbone and metric, we perform a paired t-test between \textsc{ABSS} and the corresponding baseline using per-prompt scores, and report the resulting $p$-values in Figure~\ref{fig:t1_ptest}. For readability, values below $10^{-3}$ are reported as \texttt{$<0.001$} (omitting additional digits), and values above $0.15$ are omitted since they do not indicate sufficient statistical significance. This analysis provides a quantitative measure of whether the observed improvements are consistent across prompts.

Overall, the improvements of \textsc{ABSS} over the \textsc{Random} baseline on \textit{DrawBench} and \textit{Pick-a-Pic} are relatively stable across backbones and metrics, with many comparisons yielding statistically significant $p$-values. In contrast, comparisons against the \textsc{Golden} baseline are largely significant and reliable but exhibit a few less consistent cases. This behavior is expected: "golden seeds" are typically obtained via substantially more compute-intensive procedures, whereas \textsc{ABSS} is training-free and incurs negligible additional computation at inference time. Therefore, achieving comparable or frequently significant performance against \textsc{Golden} provides strong evidence that \textsc{ABSS} is a reliable and cost-effective alternative.

Notably, \textsc{ABSS} is particularly effective on \textit{InitNO}. Across all \textit{InitNO} entries in Figure~\ref{fig:t1_ptest} (12 rows $\times$ 4 metrics = 48 tests), 41/48 $p$-values are below 0.10 (and 33/48 are below 0.05), indicating that the gains are highly consistent across prompts. Additionally, the newest DiT backbones (SD~3.0 and FLUX) show strong performance across the three datasets, with FLUX in particular exhibiting a high fraction of significant comparisons, further supporting the robustness of \textsc{ABSS} on modern diffusion and flow backbones.

\paragraph{Ranking agreement with HPS.}
\looseness-1To further validate \textsc{ABSS} based on the observations in Figure~\ref{fig:observation}, we compare the \textsc{ABSS} seed-ranking list with the ground truth HPS ranking using (i) the overlap rate—the proportion of seeds common to both lists—and (ii) Normalized Discounted Cumulative Gain (NDCG)~\citep{JarvelinKekalainen2002}, which assesses how closely an ordering matches the ideal one while applying position-based discounts. \textsc{ABSS} achieves an NDCG above 0.9 and an overlap rate of 0.68, indicating that it effectively prioritizes valuable seeds and closely approximates the ground truth ranking.

\begin{figure}[h]
  \centering
  \vspace{-2mm}
  \includegraphics[width=0.48\textwidth]{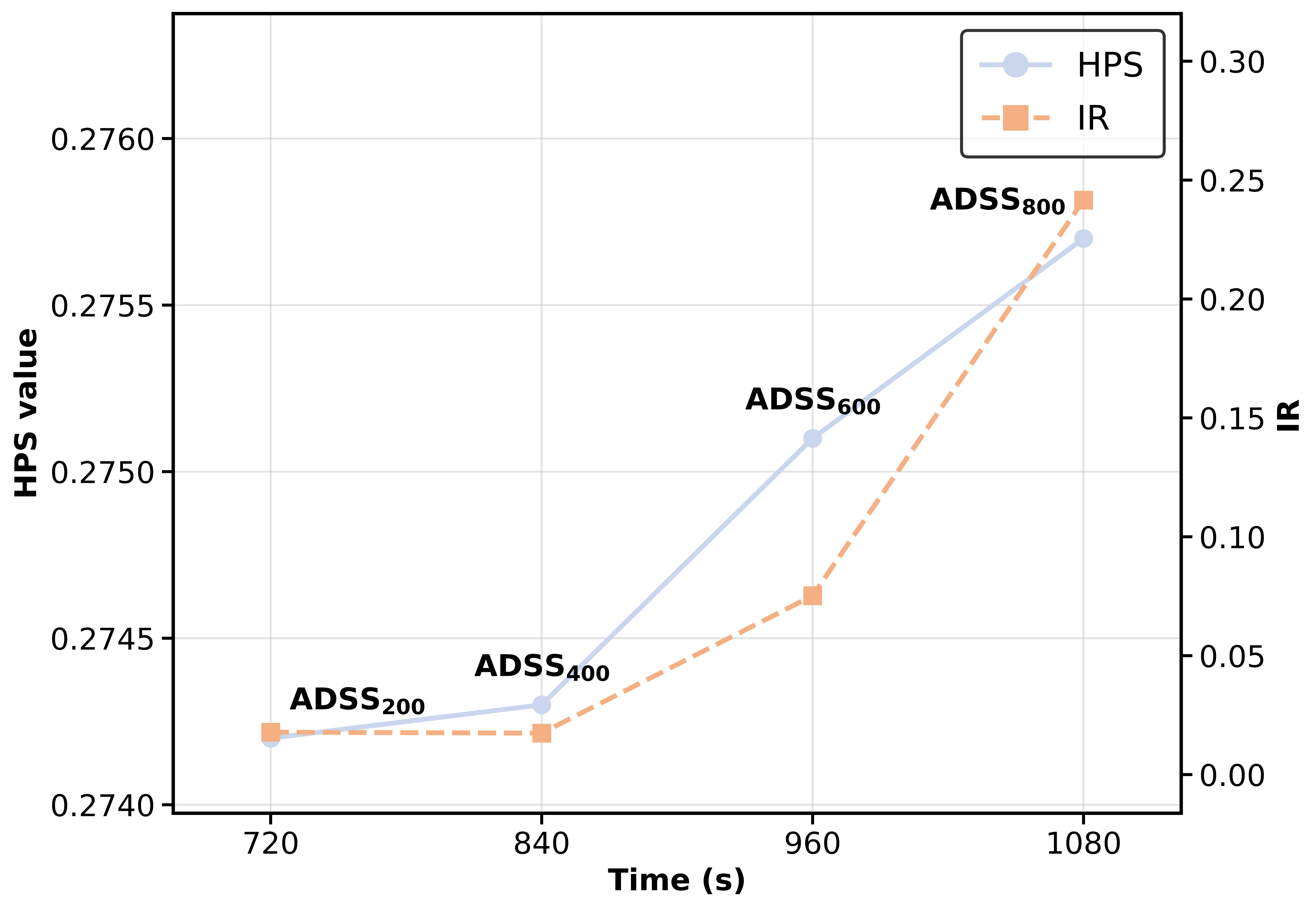}
  \vspace{-2mm}
  \caption{Performance and running time of \textsc{ABSS} on \textit{InitNO} with different timesteps.}
  \label{fig:metric}
  \vspace{-3mm}
\end{figure}

\paragraph{Effect of evaluation timestep \(t\).}
\looseness-1We also investigate \textsc{ABSS} at different timesteps and observe that performance metrics generally improve as denoising progresses. As illustrated in Figure~\ref{fig:metric}, HPS increases from 0.2742 to 0.2757 and IR from 0.0178 to 0.2416 when \textsc{ABSS} is evaluated at timesteps ranging from 200 to 800. This trend aligns with intuition: later denoising steps provide stronger signals for seed selection but incur higher computational costs. By selecting timestep $t = 200$, \textsc{ABSS} achieves a favorable trade-off between performance and efficiency, enabling practical early-stage seed selection.

\subsection{Additional Ablations on Token Types for Q3}
\label{sec:supp_q3}

\paragraph{Experimental setting.}
We conduct this ablation on \textit{DrawBench} using SD~1.4. \textit{DrawBench} prompts typically contain richer and more diverse modifiers (e.g., adjectives) and actions (verbs) than the other benchmarks, making it a suitable testbed for studying whether \textsc{ABSS} should score only core subject tokens or alternative token types. We follow the same generation and evaluation settings as in Appendix~\ref{sec:supp_table1_setting}, except that we use a larger seed pool ($N{=}100$) and retain top-$k{=}50$ seeds per prompt to better reveal the general trend and reduce randomness. In Table~\ref{table:q3t}, $\textsc{ABSS}_{\text{adj}}$ denotes scoring adjective tokens only, $\textsc{ABSS}_{\text{verb}}$ denotes scoring verb tokens only, and $\textsc{ABSS}_{\text{core}}$ denotes scoring core tokens (our default). Metrics are averaged over the selected top-$k$ generations.

\paragraph{Results.}
As shown in Table~\ref{table:q3t}, core-token scoring achieves the best overall performance across metrics. Compared with scoring adjectives or verbs alone, $\textsc{ABSS}_{\text{core}}$ consistently improves HPS and IR (and also yields slightly better PickScore and CLIP). This supports our main hypothesis that early-stage attention to the dominant subjects provides the most reliable signal for seed quality, whereas focusing on modifiers or actions can over-emphasize secondary attributes and context, leading to weaker semantic grounding of the primary objects.

\begin{table}[h]
\centering
\captionof{table}{Performance of \textsc{ABSS} with different token types on \textit{DrawBench} using SD~1.4.}
\label{table:q3t}
\resizebox{0.48\linewidth}{!}{%
\begin{tabular}{l|cccc}
  \toprule
  Method & HPS $\uparrow$ & IR $\uparrow$ & PickScore $\uparrow$ & CLIP $\uparrow$ \\
  \midrule
  $\textsc{ABSS}_{\text{adj}}$  & 0.2440 & -0.2374 & 20.8490 & 0.2617 \\
  $\textsc{ABSS}_{\text{verb}}$ & 0.2438 & -0.2272 & 20.8345 & 0.2609 \\
  $\textsc{ABSS}_{\text{core}}$ & \textbf{0.2473} & \textbf{-0.2195} & \textbf{20.8702} & \textbf{0.2619} \\
  \bottomrule
\end{tabular}}
\end{table}

\subsection{Additional Qualitative Results on Earlier SD Backbones.}
\label{sec:results_q1_images}

\begin{figure*}[h]
    \centering\vspace{-2mm}
    \includegraphics[width=\linewidth, height=0.66\textheight, keepaspectratio]{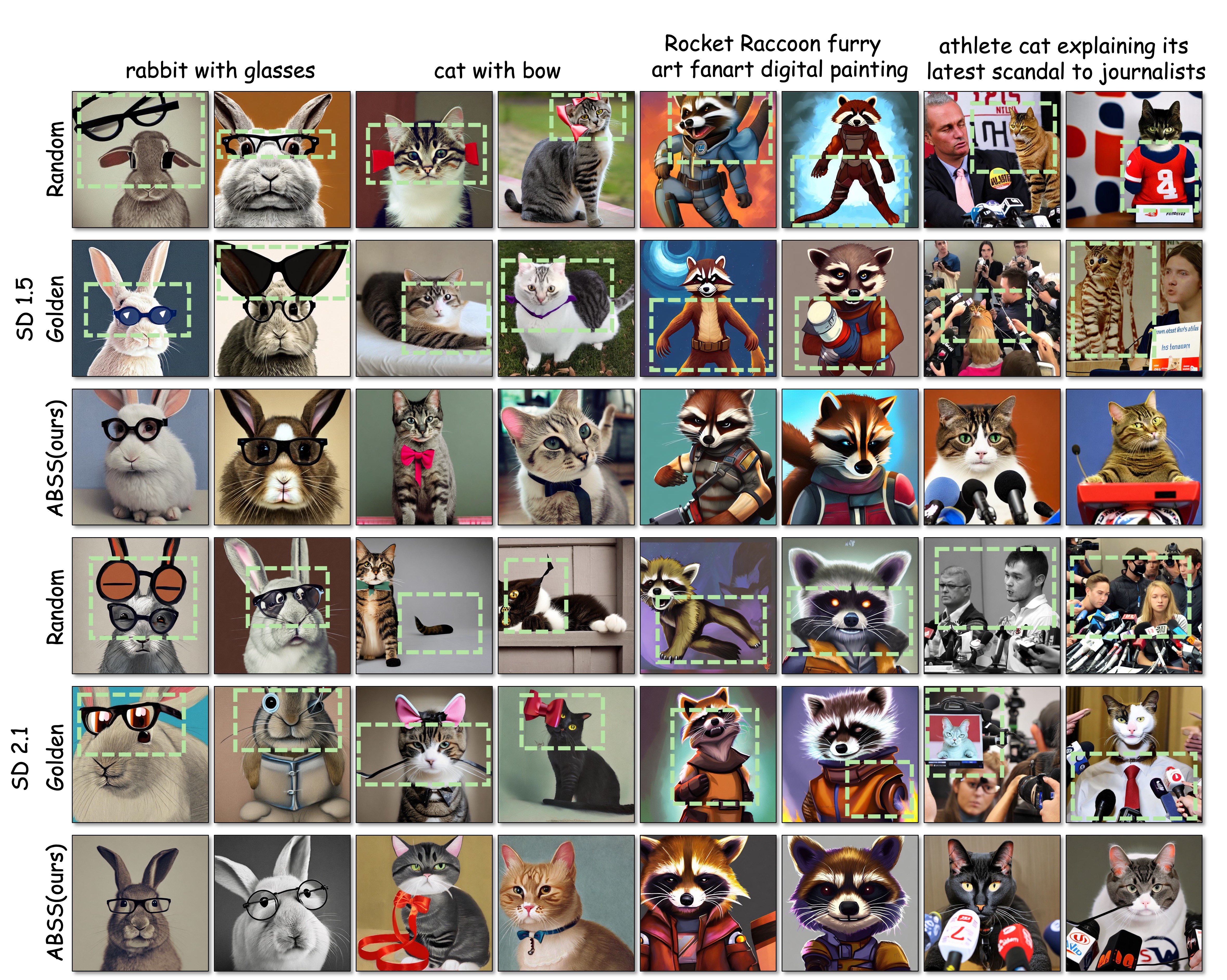}\vspace{-1mm}
    \caption{Additional qualitative comparisons on earlier Stable Diffusion backbones. Images in the same column are generated from the same prompt using \textsc{Random}, \textsc{Golden}, and \textsc{ABSS}. Pink dashed rectangles highlight problematic generated regions.}
    \vspace{-3mm}
    \label{fig:Q1_more}
\end{figure*}
We provide additional qualitative comparisons between \textsc{ABSS}, \textsc{Random}, and \textsc{Golden} on earlier Stable Diffusion backbones. These examples complement the main-paper visual results on Hunyuan-DiT and show that \textsc{ABSS} can reduce common generation errors across different model versions. Problematic regions are highlighted with orange dashed rectangles.

\end{document}